\documentclass[final,1p,times]{elsarticle}
\usepackage{amssymb}
\usepackage{amsmath}
\usepackage{multirow}
\usepackage{bbding}
\usepackage{graphicx}
\usepackage{colortbl}
\usepackage{pifont}
\usepackage[table]{xcolor} 
\usepackage{booktabs}
\usepackage[breaklinks=true,colorlinks,bookmarks=True]{hyperref}

\begin{document}
\begin{frontmatter}

\title{No Re-Train, More Gain: Upgrading Backbones with Diffusion model for Pixel-Wise and Weakly-Supervised Few-Shot Segmentation}
\author[label1]{Shuai Chen}
\author[label1]{Fanman Meng\corref{mycorrespondingauthor}}
\cortext[mycorrespondingauthor]{Corresponding author}
\ead{fmmeng@uestc.edu.cn}
\author[label1]{Chenhao Wu}
\author[label1]{Haoran Wei}
\author[label1]{Runtong Zhang}
\author[label1]{Qingbo Wu}
\author[label1]{Linfeng Xu}
\author[label1]{Hongliang Li}
\affiliation[label1]{organization={University of Electronic Science and Technology of China}, city={Chengdu}, postcode={611731}, country={China}}

\begin{abstract}
  Few-Shot Segmentation (FSS) aims to segment novel classes using only a few annotated images. Despite considerable progress under pixel-wise support annotation, current FSS methods still face three issues: the inflexibility of backbone upgrade without re-training, the inability to uniformly handle various types of annotations (e.g., scribble, bounding box, mask, and text), and the difficulty in accommodating different annotation quantity. To address these issues simultaneously, we propose DiffUp, a novel framework that conceptualizes the FSS task as a conditional generative problem using a diffusion process. For the first issue, we introduce a backbone-agnostic feature transformation module that converts different segmentation cues into unified coarse priors, facilitating seamless backbone upgrade without re-training. For the second issue, due to the varying granularity of transformed priors from diverse annotation types (scribble, bounding box, mask, and text), we conceptualize these multi-granular transformed priors as analogous to noisy intermediates at different steps of a diffusion model. This is implemented via a self-conditioned modulation block coupled with a dual-level quality modulation branch. For the third issue, we incorporate an uncertainty-aware information fusion module to harmonize the variability across zero-shot, one-shot, and many-shot scenarios. Evaluated through rigorous benchmarks, DiffUp significantly outperforms existing FSS models in terms of flexibility and accuracy.
\end{abstract}

\begin{keyword}
  Few-shot segmentation \sep diffusion model \sep backbone upgrade \sep multi-granularity
\end{keyword}

\end{frontmatter}
\section{Introduction}
\label{intro}
Object segmentation serves as a foundational task in computer vision, with applications spanning from autonomous driving to medical imaging. Despite the considerable success of fully supervised segmentation method~\cite{long2015fully}, its reliance on extensive annotated datasets imposes substantial limitations. The acquisition and annotation of such datasets are both time-consuming and labor-intensive. Furthermore, these methods inherently lack the capability to generalize to unseen categories. In this context, Few-Shot Segmentation (FSS) paradigm has emerged as a promising alternative, leveraging a few annotated examples to facilitate rapid adaptation to new categories.

\begin{figure}[htbp]
  \centering
  \includegraphics[width=1. \linewidth]{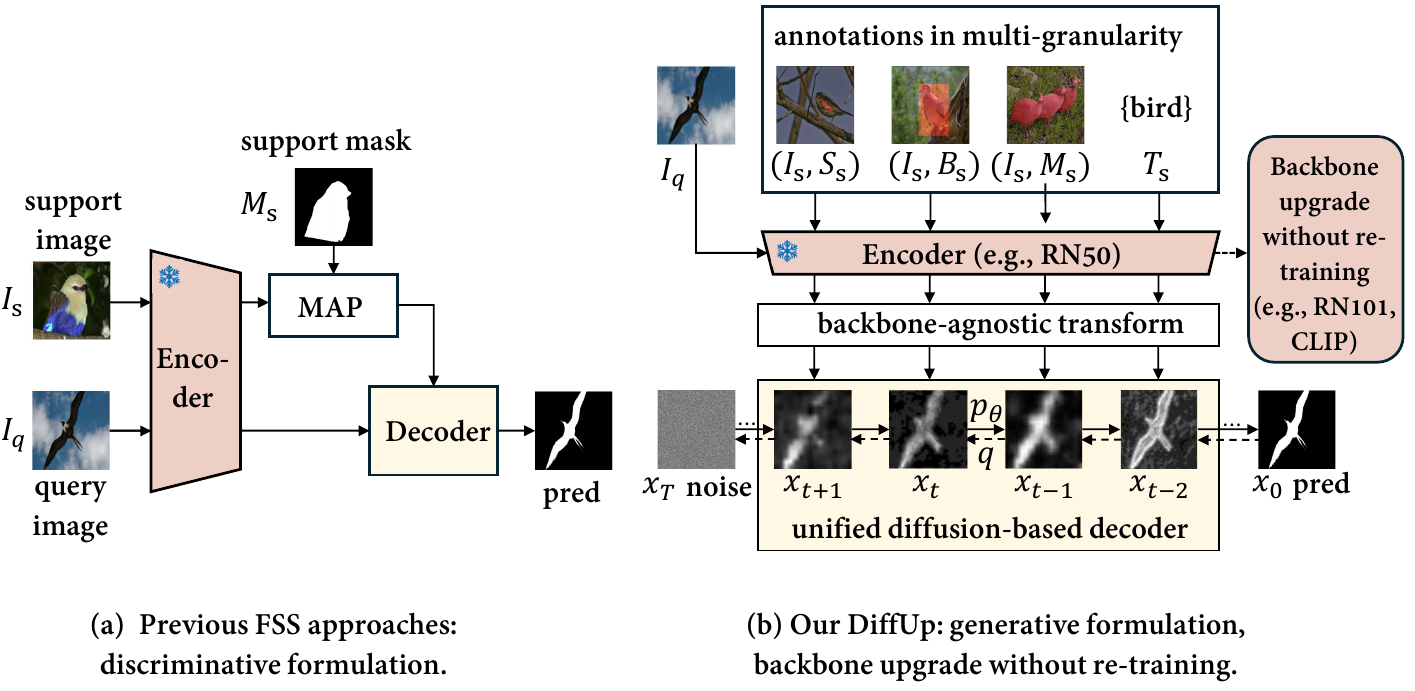}
  \caption{Comparison of approaches: (a) Traditional FSS methods use discriminative formulations, limiting backbone upgrades without retraining and struggling with diverse annotation types/quantities. (b) Our DiffUp adopts a generative approach, treating varying segmentation cues as diffusion process intermediates, enabling flexible backbone upgrades and supporting multiple annotation forms (scribbles, bounding boxes, masks, text) in various quantities.}
  \label{fig:motivation}
\end{figure}

Current FSS methods, whether prototype-based~\cite{wang2019panet,tian2020prior,liu2020part} or matching-based approaches~\cite{shi2022dense,min2021hypercorrelation}, have achieved significant progress under pixel-level annotation. However, these methods still face several unresolved challenges that hinder their practical deployment. These challenges include: (I) the unexplored aspect of \textbf{backbone upgrade without re-training} in the FSS field, as depicted in the top of Fig.~\ref{fig:motivation}, FSS models typically incorporate a pre-trained, frozen backbone, coupled with a backbone-sensitive decoder, necessitating re-training for different backbones. Such a workflow is both inflexible and inefficient, particularly when re-training is hindered by the unavailability of previous training data due to privacy concerns~\cite{gong2024continual}. An ideal solution would permit backbone upgrade without re-training, encompassing scale-up (e.g., RN50 to RN101~\cite{DBLP:conf/cvpr/HeZRS16}), architectural shifts (e.g., CNNs~\cite{DBLP:conf/cvpr/HeZRS16} to ViTs~\cite{DBLP:conf/iclr/DosovitskiyB0WZ21}), and pre-training adjustments (e.g., ImageNet~\cite{DBLP:conf/cvpr/DengDSLL009} to CLIP~\cite{radford2021learning}). (II) the inability to \textbf{uniformly handle various weak and pixel annotations} (e.g., bounding box, scribble, text and mask). This necessitates the development of specialized networks tailored to each annotation type, reducing the system's training and deployment flexibility. (III) the challenge of bridging the gap among \textbf{varied quantities} (e.g., zero-shot, one-shot, and many-shot) presents another complexity. Both Zero-Shot Segmentation (ZSS) and FSS aim to segment new categories, but are historically modeled separately. This separation, primarily due to ZSS's lack of visual samples, limits the efficacy in scenarios where unseen and minimally represented categories must be segmented within a unified framework. Additionally, transitioning from one-shot to many-shot segmentation increases complexity by requiring multiple inferences.

Most earlier advancements~\cite{wang2019panet,tian2020prior,liu2020part,shi2022dense,min2021hypercorrelation} in FSS primarily focus on pixel-wise annotations, often overlooking the above three challenges. While some pioneering works have partially explored these issues, a comprehensive solution is still lacking. For instance, regarding annotation diversity, IMR-HSNet~\cite{wang2023iterative}, MLFWS~\cite{MLFWS}, and PartSeg~\cite{PartSeg} have initiated exploration into weakly supervised scenarios, covering image-level, bounding boxes, and textual annotations. In terms of quantity variability, CLIPSeg~\cite{luddecke2022image} seeks to merge zero-shot and one-shot scenarios through mixup techniques, yet it lacks comprehensive modeling for many-shot scenarios. Furthermore, the FSS field has not yet investigated the potential for updating backbone without re-training. 

This paper introduces DiffUp, a novel FSS framework that simultaneously addresses these three challenges in a unified manner. Specifically, regarding \textbf{backbone upgrade without re-training}, our research identifies that variations in feature dimension and subspace across different backbones render direct replacements impractical. To overcome this, we propose a Backbone-Agnostic Feature Transformation module (BAFT). This module converts varied segmentation cues (e.g., scribble, bounding box, mask and text) into coarse priors. Being backbone-agnostic, the coarse priors facilitate straightforward backbone upgrade without re-training.

Regarding \textbf{annotation diversity}, while the BAFT module processes diverse support annotations into the same format in coarse priors, it does not consider the inherent multi-granularity of these priors (i.e., the priors under diverse segmentation cues are different), thus leading to sub-optimal outcomes. We observe that during the forward stage of a diffusion model~\cite{rombach2022high}, the noise content gradually intensifies. This parallels the process of changing priors from fine-grained segmentation cues (e.g., mask) to those that are coarse-grained (e.g., scribble). Such insight suggests formulating the FSS task as a conditional diffusion process. To ensure that the multi-granular priors effectively guide the diffusion process, we further introduce a Unified Quality-aware Diffusion-based Decoder (UQDD), which includes a Self-Conditioned Modulation (SCM) module coupled with a Dual-level Quality Modulation (DQM) branch.

Regarding \textbf{varied quantities} of support samples, contrary to existing methods that rely on multiple inferences for ensembles, our framework views varying quantities of support samples as different perspectives on the query. This perception allows for the introduction of an Uncertainty-Aware Prior Fusion module (UAPF), equipped with mean-covariance representations. This module proficiently addresses the any-shot challenge, from zero-shot to many-shot scenarios. The contributions of the proposed DiffUp are as follows:

\begin{enumerate}
    \item Backbone upgrade without re-training: to the best of our knowledge, this is the first attempt to introduce the concept of backbone upgrade without re-training in the FSS field. This is achieved by the Backbone-Agnostic Feature Transformation (BAFT) module.
    \item Generative problem formulation: we conceptualize FSS as a conditional diffusion process, where multi-granular priors are intricately linked to the noise intermediates in the diffusion process. This is realized by the Unified Quality-Aware Diffusion-based Decoder (UQDD).
    \item Unified model architecture: we propose a model architecture that can effectively handle various annotation types (e.g., scribble, bounding box, mask, and text) as well as different quantities (ranging from zero-shot and one-shot to many-shot scenarios).
    \item Superior performance: our model achieves enhanced performance through rigorous benchmarks, achieving 74.9 and 54.6 mIoU on $\text{PASCAL-}5^i$ and $\text{COCO-}20^i$ datasets, respectively. 
\end{enumerate}

\section{Related Work}
\subsection{Few-shot Segmentation}
Current FSS methods could be broadly categorized into prototype-based and matching-based approaches. Prototype-based methods involved condensing the support features into class-specific prototypes. Various strategies have been proposed to construct these prototypes: some methods generated a single global foreground prototype through masked average pooling~\cite{ wang2019panet}, while others created multiple foreground-background prototypes using learnable modules~\cite{wang2024adaptive, wang2022remember}, prototype interaction~\cite{bao2024relevant}, and clustering~\cite{li2024label}. To address the spatial information loss inherent in prototype-based methods, matching-based methods had been developed to establish dense pixel-to-pixel correspondences between support and query images. These methods employed a range of techniques, including mask assembly~\cite{ shi2022dense}, hypercorrelation squeeze~\cite{min2021hypercorrelation}, and maximizing support-set information~\cite{moon2023msi}. Moreover, the advancements in large-scale models have inspired several approaches to integrate these priors into FSS. For example, some methods incorporated CLIP~\cite{radford2021learning} priors into FSS~\cite{luddecke2022image, wang2023iterative, guo2023clip}, while others explored text-to-image priors~\cite{tan2023diffss}. In contrast to these approaches, our method conceives the FSS task as a conditional generative formulation using a diffusion process, ensuring consistency across varying granularity of diverse annotations and enabling the upgrading of backbones.
 
\subsection{Backbone Upgrade}
The hot upgrading of backbones has recently emerged as a significant research focus, aiming to ensure feature compatibility between new and existing models. This advancement facilitates the accuracy-efficiency trade-off and mitigates data limitations. However, this process presents challenges due to inconsistencies in feature spaces and dimensions. To address these issues, TaCA~\cite{zhang2023taca} introduced a task-agnostic compatible adapter, demonstrating effectiveness in tasks such as visual question answering. Similarly, X-Adapter~\cite{ran2024x} showcased universal compatibility with plugins for upgraded diffusion models in the field of controllable image generation. SN-Net~\cite{pan2023stitchable} employed trainable stitching layers to interpolate and upgrade the backbones. Furthermore, ESTA~\cite{he2024efficient} optimized models under varying resource constraints via efficient fine-tuning and streamlined deployment. Despite these advancements, these methods encounter two major obstacles: (i) inability to generalize to unseen backbones during training, (ii) inability to generalize to unseen categories. In response, this paper introduces, for the first time in the context of few-shot segmentation, the practical deployment issue of hot upgrading backbones, while also ensuring generalization to unseen categories with diverse annotations.

\subsection{Generative-based Segmentation}

Generative models have emerged as powerful tools for various segmentation tasks, with approaches broadly categorized into GAN-based~\cite{tritrong2021repurposing,zhang2021datasetgan,han2022leveraging} and diffusion-based segmentation methods~\cite{prabhudesai2024test,qi2024unigs,amit2021segdiff,wu2024medsegdiffv2}. In the GAN-based direction, several pioneering works have demonstrated the effectiveness of utilizing pre-trained GANs' rich semantic knowledge. DatasetGAN~\cite{zhang2021datasetgan} pioneered this approach by augmenting pre-trained GANs with a dedicated labeling branch, enabling automated dataset annotation. Subsequently, RepurposeGAN~\cite{tritrong2021repurposing} and PFTGAN~\cite{han2022leveraging} revealed that GANs optimized for image synthesis naturally encode features conducive to few-shot part segmentation. 

Diffusion models, initially popular in image generation~\cite{rombach2022high}, have emerged as another promising direction for segmentation tasks, owing to their training stability. These approaches manifest in two primary categories. The first category leverages pre-trained stable-diffusion models~\cite{rombach2022high}. For instance, Diffusion-TTA~\cite{prabhudesai2024test} applied generative feedback for test-time adaptation in segmentation tasks. UniGS~\cite{qi2024unigs} unified image generation and segmentation within the color space. The second category directly employs the diffusion process without text-to-image knowledge, treating segmentation as a conditional generation problem~\cite{amit2021segdiff}. In medical image segmentation, MedSegDiffv2~\cite{wu2024medsegdiffv2} utilized spectrum-space transformation to guide the diffusion process effectively. Our proposed DiffUp fits into the second category of diffusion-based approaches, uniquely framing the few-shot segmentation task as a generative problem via diffusion. Unlike previous diffusion-based segmentation approaches that conditioned on backbone-sensitive features, DiffUp uses backbone-agnostic coarse priors. These coarse priors in diverse granularities were tightly linked to the noise intermediates in the diffusion process.

\section{Methodology}

\subsection{Problem Setup}

Few-shot segmentation aims to segment novel classes with only a limited number of labeled images. To achieve this, the model trained on the base categories $\boldsymbol{C}_{b}$ of base dataset $\mathcal{D}_{b}$, must possess the ability to provide reliable inference on the novel categories $\boldsymbol{C}_{n}$ of novel dataset $\mathcal{D}_{n}$, noting that $\boldsymbol{C}_{b} \cap \boldsymbol{C}_{n}=\emptyset$. The episodic sampling~\cite{tian2020prior,min2021hypercorrelation,shi2022dense} is utilized during both the training and testing phases, where each episode task $\mathcal{T}$ comprises a support set $\mathcal{S}$ and a query set $\mathcal{Q}$ in the $N$-way $K$-shot manner, i.e.,
\begin{equation}
\mathcal{T} = (\mathcal{S}, \mathcal{Q}), 
\end{equation}
where $\mathcal{S}=\left\{\left(I_{s}, M_{s}\right)\right\}_{i=1}^{N \times K}$, and $\mathcal{Q}=\left\{\left(I_{q}, M_{q}\right)\right\}_{i=1}^{N}$. Here, $N$ represents the number of classes sampled per episode, and $K$ is the number of labeled images available per class, typically 1 or 5. The support set $\mathcal{S}$ contains a few of labeled images $(I_{s}, M_{s})$ for each class, and the query set $\mathcal{Q}$ includes images $(I_{q}, M_{q})$ to be segmented, with $I_{s}, I_{q} \in \mathbb{R}^{H \times W \times 3}$ and $M_{s}, M_{q} \in \mathbb{R}^{H \times W}$ representing the images and ground-truth masks for support and query sets respectively.

In this work, we define the upgrading problem within a series of models $\mathcal{G}=\{g_i\}_{i=1}^{n} = \{(Enc_{i}, Dec_{i})\}_{i=1}^{n}$, each differentiated by its backbone $Enc_{i}$, including backbone scale-up (e.g., RN50 to RN101~\cite{DBLP:conf/cvpr/HeZRS16}), backbone architecture transition (e.g., CNN~\cite{DBLP:conf/cvpr/HeZRS16} to ViT~\cite{DBLP:conf/iclr/DosovitskiyB0WZ21}), and variations in pre-training strategy (e.g., ImageNet~\cite{DBLP:conf/cvpr/DengDSLL009} to CLIP~\cite{radford2021learning}). Given the disparate feature spaces and dimensions across encoders, the conventional decoder $Dec_{i}$ tailored to $Enc_{i}$ cannot be directly stitched to another $Enc_{j}$. Our focus is on developing a strategy that allows for cross-backbone generalization without re-training. It's formulated as: 

\begin{equation}
    \theta^{*} = \arg\min_{\theta} \mathcal{L}\left(Dec(Enc_{i}(\mathcal{T}))\right), \quad \mathcal{T} \in \mathcal{D}_{b},
\end{equation}
where $\mathcal{L}$ is the training loss function, $\theta^{*}$ represents the optimized parameters of the universal decoder trained with a specific encoder $Enc_i$. The effectiveness of $\theta^{*}$ is then evaluated on unseen encoders $Enc_j$ on novel dataset $\mathcal {D}_{n}$ without any re-training through the performance metric $\mathcal{P}$ as:
\begin{equation}
    \mathcal{P}\left(Dec (Enc_{j}(\mathcal{T})); \theta^{*}\right), \quad \forall j \neq i, \quad \mathcal{T} \in \mathcal{D}_{n}.
\end{equation}

Moreover, we expand the FSS framework to accommodate a broader range of annotation types and to include ZSS tasks. The annotations $A_{s}$ extensively include mask $M_{s}$, bounding box $B_{s}$, and scribble $S_{s}$. We also extend it to ZSS task by allowing $K=0$ (no support images are available), and introducing text embedding $T_{s}$. Then the expanded support set $\hat{\mathcal{S}}$ is formulated as:
\begin{equation}
\hat{\mathcal{S}} = \{\left\{\left(I_{s}, A_{s}\right)\right\}_{i=1}^{N \times K},  T_{s}\}, A_{s} \in \{M_{s}, B_{s}, S_{s} \}.
\end{equation}

\subsection{Preliminary of Diffusion Models}

Diffusion models~\cite{ho2020denoising} represent a promising category of generative models that exhibit exceptional stability in data generation and transformation. These models integrate both a forward process, denoted as $q(x_t|x_{t-1})$, and a reverse process, represented by $p(x_{t-1}|x_{t})$, to synthesize new samples. The forward process $q(x_t|x_{t-1})$ involves progressively introducing Gaussian noise $\epsilon$ to transform a clean sample $x_0$ into its noisy counterparts $[x_1,..., x_t,..., x_T]$ as follows:

\begin{align}
    q(x_t|x_{t-1}) &= \mathcal{N}(x_t; \sqrt{1-\beta_t} x_{t-1}, \beta_t \mathbf{I}), \\
    q(x_t|x_0) &= \mathcal{N}(x_t;\sqrt{\bar{\alpha}_{t}} x_0,\left(1-\bar{\alpha}_{t}\right) \mathbf{I}),
\end{align}
where $\alpha_{t}$ represents the noise level at step $t$, controlled by noise schedule parameters $\beta_{t}$, i.e., $\alpha_{t}=1-\beta_{t}$, $\bar{\alpha}_{t}=\prod_{s=1}^{t} \alpha_{s}$. Conversely, the reverse process $p(x_{t-1}|x_{t})$ starts with  an initial Gaussian noise sample $x_T $, and progressively reconstructs the original data  $x_0$ by the order $[x_{T-1}, ..., x_t, x_{t-1}, ..., x_0]$ as:
\begin{equation}
    p\left(x_{t-1}|x_{t}\right)=\mathcal{N}\left(x_{t-1} ; \boldsymbol{\mu}_{\theta}\left(x_{t}, t\right), \mathbf{\Sigma}_{\theta}\left(x_{t}, t\right)\right),
\end{equation}
where $\boldsymbol{\mu}_{\theta}\left(x_{t}, t\right)$ denotes the learnable expectation, and $\mathbf{\Sigma}_{\theta}\left(x_{t}, t\right)$ represents the variance. Training conditional DDPMs can be efficiently approximated by optimizing a network $\theta$ to predict the noise $\epsilon$ added to a clean sample $x_0$:

\begin{align}
    \mathcal{L}_{diff} &= \mathbb{E}_{x_0,t,\epsilon,c} \left[\|\epsilon - \epsilon_\theta(x_t, t,c)\|^2\right] \nonumber \\
    &= \mathbb{E}_{x_0,t,\epsilon,c} \left[\|\epsilon - \epsilon_\theta(\sqrt{\bar{\alpha_t}}x_0 + \sqrt{1 - \bar{\alpha_t}}\epsilon, t,c)\|^2\right],
\end{align}
where $c$ is the condition to guide the diffusion process.

\subsection{Proposed Method: DiffUp}
\begin{figure}[t]
    \centering
    \includegraphics[width= 1.0\linewidth]{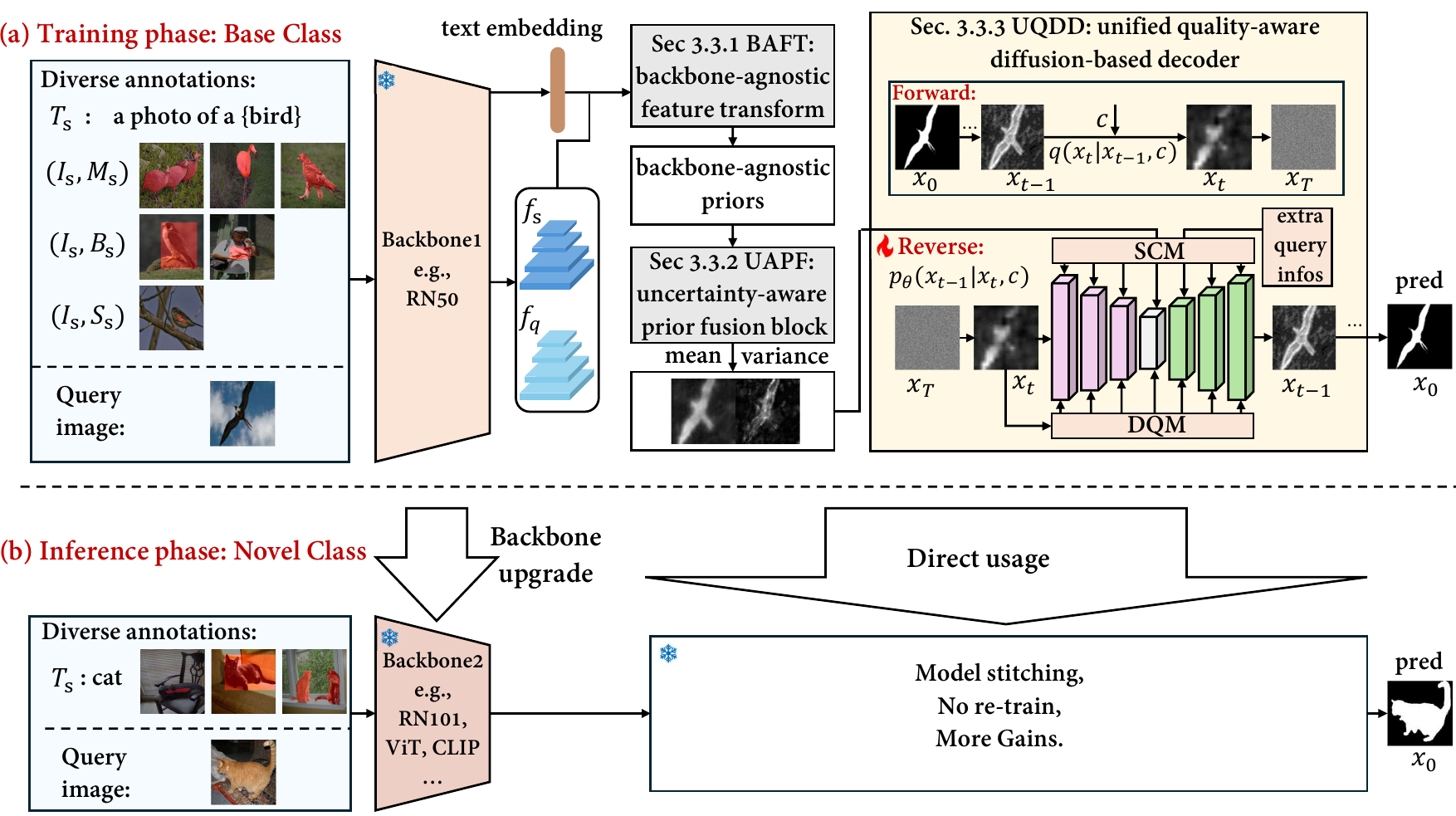}
    \caption{Overview of DiffUp. (a) Training phase: Multi-scale features from base classes are extracted via a frozen RN50 backbone under various annotations (scribble, bounding-box, pixel-level, textual). Features undergo BAFT projection to a universal space (Section~\ref{subsubsection:BAFT}), enabling seamless backbone upgrades. UAPF refines features by fusing segmentation cues with varying certainties (Section~\ref{subsubsection:UAPF}). These refined priors condition a quality-aware diffusion decoder (Section~\ref{subsubsection:UQDD}) to generate precise segmentations from Gaussian noise. (b) Inference phase: The system accommodates backbone upgrades through scaling (RN50 to RN101), architecture transitions (CNNs to ViTs), and different pre-training strategies (ImageNet to CLIP), ensuring robust performance on novel classes.}
    \label{fig:framework}
  \end{figure}

We propose DiffUp, a novel FSS method addressing three challenges: backbone upgrade without re-training, handling diverse annotation types, and accommodating varying annotation quantities. As illustrated in Fig.~\ref{fig:framework}, DiffUp consists of three key components: a Backbone-Agnostic Feature Transform (BAFT) block (Section \ref{subsubsection:BAFT}), an Uncertainty-Aware Prior Fusion (UAPF) block (Section \ref{subsubsection:UAPF}), and a Unified Quality-aware Diffusion-based Decoder (UQDD) (Section \ref{subsubsection:UQDD}). These modules convert heterogeneous segmentation cues into universal priors, incorporate uncertainty, and process multi-granularity via diffusion, respectively.

\begin{figure}
    \centering
    \includegraphics[width= 1.0\linewidth]{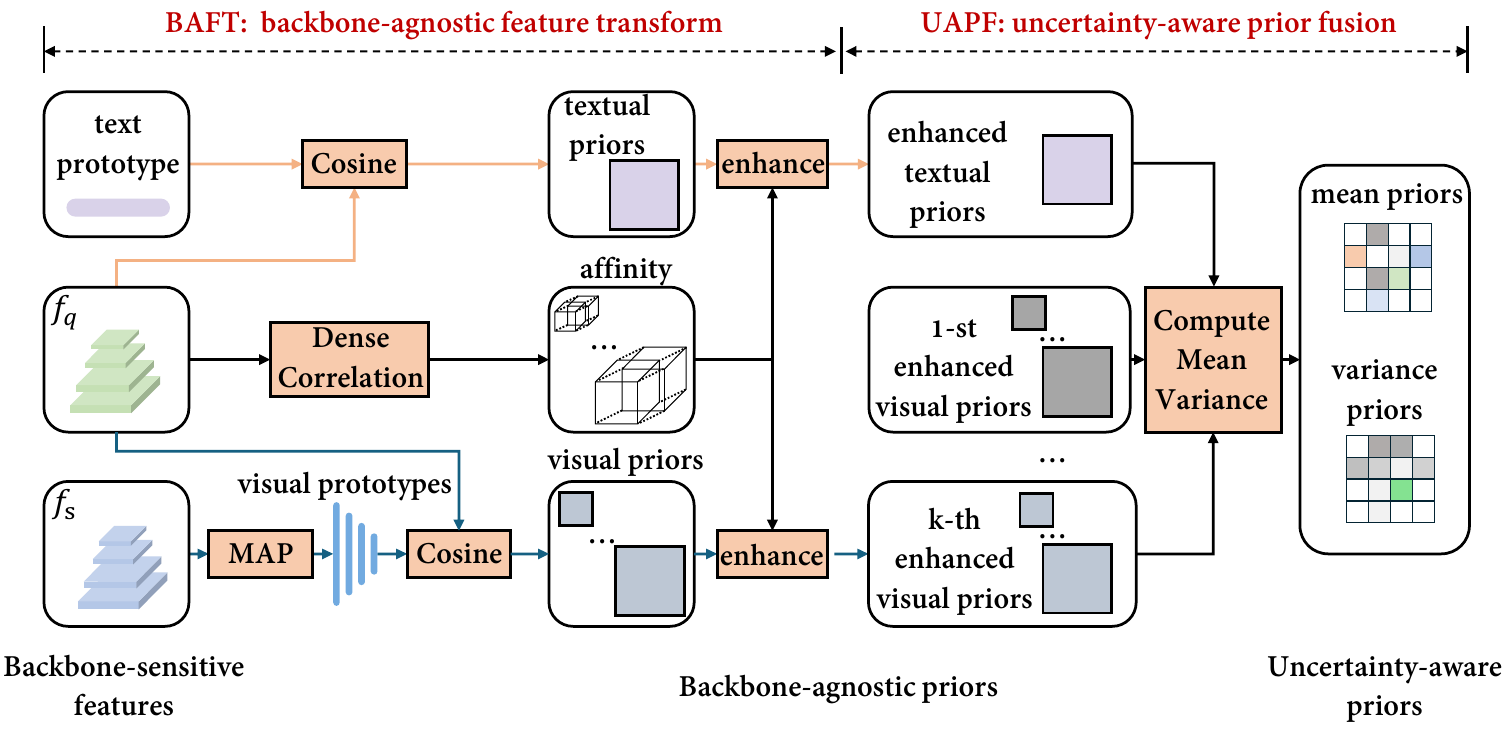}
    \caption{The Backbone-Agnostic Feature Transform (BAFT) block converts diverse segmentation cues into unified, backbone-agnostic priors. And the Uncertainty-Aware Prior Fusion (UAPF) block fuses these priors, handling varying annotation quantities and incorporating uncertainty.}
    \label{fig:BAFT-UAFT}
\end{figure}

\subsubsection{Backbone-Agnostic Feature Transform (BAFT)}
\label{subsubsection:BAFT}
The heterogeneity of feature spaces and dimensions across diverse backbones impedes the capacity of extant FSS models to upgrade backbones without re-training. To address this constraint, we introduce the Backbone-Agnostic Feature Transform (BAFT) block, which transforms backbone-specific features into a universal, backbone-agnostic representation that can generalize to novel backbones on novel classes.

The pipeline of BAFT is illustrated in Fig.~\ref{fig:BAFT-UAFT}. For the backbone $Enc_i$ available at the current training stage, multi-scale features are first extracted for both support and query images. At layer $l$ of a backbone, we obtain feature set $f_{Enc_i}^l = ({f^l_{s,Enc_i}, f^l_{q,Enc_i}})$, consisting of support feature $f^l_{s,Enc_i} \in \mathbb{R}^{h_s^l \times w_s^l \times d_i}$ and query feature $f^l_{q,Enc_i} \in \mathbb{R}^{h_q^l \times w_q^l \times d_i}$, with $d_i$ denoting the feature dimension of current backbone $Enc_i$. To accommodate diverse visual annotations $A_{s}$ (e.g., scribble, bounding box, mask), we uniformly treat them as pixel-wise annotations and apply masked average pooling (MAP) on the backbone-sensitive support feature $f^l_{s,Enc_i}$ to derive visual prototypes $proto_v^l$. For text annotations, we employ a text encoder to extract textual prototypes $proto_t$. For brevity, the layer index $l$ is omitted hereafter.

Subsequently, these backbone-sensitive features/prototypes are transformed into backbone-agnostic priors $p$ through transformation function $\Phi$: 
\begin{equation}
p = \Phi(f_{s,Enc_i}, A_{s},f_{q,Enc_i}),
\end{equation}
where $A_{s}$ is the diverse annotations of support image. Given that cosine similarity inherently provides both dimension unification and amplitude normalization properties, we adopt the cosine function as a simple yet effective implementation of $\Phi$. This design choice naturally projects features from different backbones into a unified probability space. We leave the exploration of other universal transformation function for future work. Specifically, we compute pairwise cosine similarities between prototypes and query features to obtain backbone-agnostic visual priors $p_v \in \mathbb{R}^{h_{q} \times w_{q}}$ and textual priors $p_t \in \mathbb{R}^{h_{q} \times w_{q}}$ as:

\begin{equation}
\label{eq:cosine_sim}
p_v = \frac{f_{q,{Enc_{i}}} \cdot {proto_v}^T}{\|f^{l}_{q,{Enc_{i}}}\| \|proto_v\|},
p_t = \frac{f_{q,{Enc_{i}}} \cdot {proto_t}^T}{\|f^{l}_{q,{Enc_{i}}}\| \|proto_t\|}.
\end{equation}

Furthermore, to exploit the self-similarity information within the query image, we introduce an affinity-enhanced strategy. Specifically, we compute the affinity matrix $A_{ff} \in \mathbb{R}^{(h_{q} \times w_{q}) \times (h_{q} \times w_{q})}$ of the query feature, where each element ${{A_{ff}}_{(i,j)}}$ measures the cosine similarity between the $i$-th and $j$-th feature vectors. The affinity-enhanced backbone-agnostic visual prior $p_v^{aff} \in \mathbb{R}^{h_{q} \times w_{q}}$ and textual prior $p_t^{aff} \in \mathbb{R}^{h_{q} \times w_{q}}$ are then obtained as:

\begin{equation}
\label{eq:aff_prior}
p_v^{aff} = A_{ff} \cdot p_{v}^{flat}, p_t^{aff} = A_{ff} \cdot p_{t}^{flat},
\end{equation}
where $p_{v}^{flat}$ and $p_{t}^{flat}$ are the flattened version of visual prior $p_v$ and textual prior $p_t$ respectively.

\subsubsection{Uncertainty-Aware Prior Fusion block (UAPF)}
\label{subsubsection:UAPF}
For the $K$-shot situation, existing FSS methods~\cite{tian2020prior,min2021hypercorrelation} often resort to multiple one-shot inference followed by ensemble. While straightforward, it fails to fully exploit the rich information contained in the varied support samples. In contrast, our framework views the varying quantities of support samples as different viewpoints on the query. This insight motivates the development of an Uncertainty-Aware Information Fusion module, empowered by mean-covariance representations. As illustrated in Fig.~\ref{fig:BAFT-UAFT}, the UAPF block operates on the unified representations obtained by BAFT block to each annotated support-query pair. For layer $l$, let $\{p_1^l, p_2^l, \ldots, p_k^l \}$ denote the $k$ transformed priors (including $p_v^{aff}$ and $p_t^{aff}$) corresponding to the $K$ support-query pairs. The UAPF block computes the mean prior $\mu^l(p)$ and the variance prior $\Sigma^l(p)$ at each layer as follows:

\begin{equation}
    \mu^l(p) = \frac{1}{K} \sum_{i=1}^K p_i^l,  \Sigma^l(p) = \frac{1}{K-1} \sum_{i=1}^K (p_i^l - \mu^l(p))^2.
\end{equation}

The variance prior $\Sigma^l(p)$ quantifies the level of variability and uncertainty associated with the fused information. In the extreme case of 1-shot learning, the UAPF block employs data augmentation techniques to expand the single support sample into multiple samples, enabling the computation of variance. Additionally, we select a fixed number of priors as extra outputs of the UAPF block. While BAFT and UAPF employ simple operations, their parameter-free design is crucial for achieving backbone flexibility and support sample adaptability.

\subsubsection{Unified Quality-aware Diffusion-based Decoder (UQDD)}
\label{subsubsection:UQDD}

Transforming coarse, backbone-agnostic priors into precise segmentation results presents a significant challenge. A simplistic approach might involve merging these priors with low-resolution features and employing an up-sampler for single-step inference. However, this would introduce backbone-sensitive features and dimension-specific up-samplers, contradicting our objective of facilitating backbone upgrade without re-training. To address this issue, we propose the Unified Quality-aware Diffusion-based Decoder (UQDD). This module leverages the iterative refinement characteristic of the diffusion process to align with the varying granularity of  priors, ranging from weak supervisions (scribble, bounding box, and text) to strong supervision with mask.

Specifically, we employ a U-Net architecture where the segmentation ground-truth $M_q$ serves as the clean sample $x_0$. The diffusion process maintains a layer-wise correspondence between the backbone and UNet: for each backbone layer $l$, its outputs are processed by BAFT and UAPF, then are directly fed into the corresponding UNet layer $l$ as conditions. This layer-wise conditional information $c^l$ integrates two components: the statistical representations of backbone-agnostic priors (mean $\mu^l(p)$ and variance $\Sigma^l(p)$), and backbone-independent query information (e.g., Canny, Sobel, and VAE features). The forward process introduces Gaussian noise to the segmentation mask $x_0$ over $T$ timesteps, producing progressively noisier masks ${x_1, x_2, ..., x_T}$. This process is formulated as a Markov chain with a fixed variance schedule $\beta_1, \beta_2, \ldots, \beta_T$:

\begin{equation}
q(x_t | x_{t-1}, c) = \mathcal{N}(x_t; \sqrt{1-\beta_t}x_{t-1} + \mu_c, \beta_t \mathbf{I}),
\end{equation}
where \(\mu_c\) is an offset determined by the condition \(c\). The conditional reverse process is formulated as:

\begin{equation}
p\left(x_{t-1} | x_t, c\right) = \mathcal{N}\left(x_{t-1} ; \boldsymbol{\mu}_{\theta}\left(x_t, t, c\right), \mathbf{\Sigma}_{\theta}\left(x_t, t\right)\right).
\end{equation}

To ensure the condition $c$ effectively guides the sampling process and to accommodate multiple levels of granularity, we implement UQDD using two key components: the proposed Self-Conditioned Modulation (SCM) block and a Dual-level Quality Modulation (DQM) branch. The latter imposes constraints through spatially-informed error maps and global IoU measurements, respectively.

\begin{figure}[htbp]
  \begin{minipage}[t]{0.43\textwidth}
      \centering
      \includegraphics[width= 1.\linewidth]{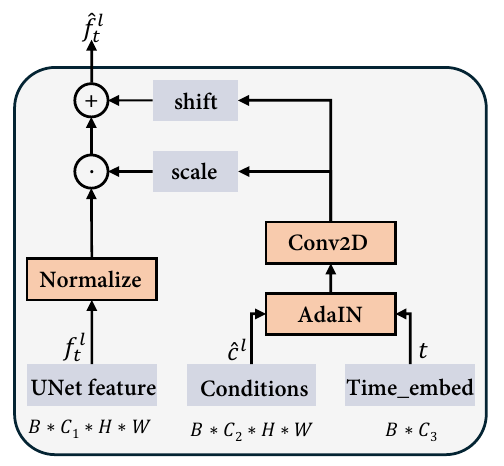}
      \caption{The pipeline of the porposed Self-Conditioned Modulation Block (SCM).}
      \label{Self-Adaptive Modulation Block (SCM)}
  \end{minipage}
  \hfill
  \begin{minipage}[t]{0.55\textwidth}
      \centering
      \includegraphics[width= 1. \linewidth]{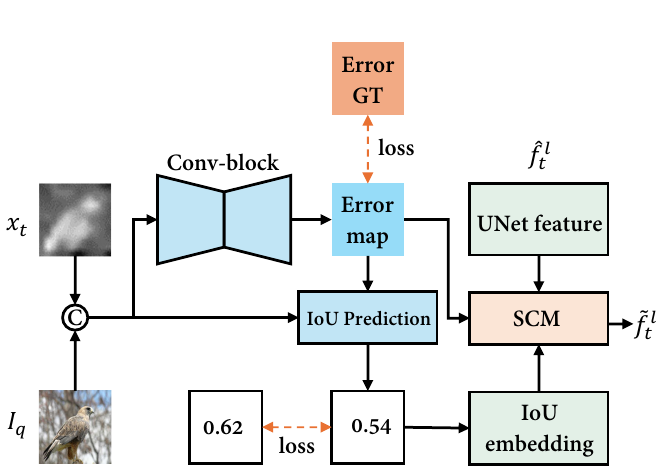}
      \caption{The pipeline of Dual-level Quality Modulation branch (DQM), including the error map level and IoU level modulations.}
      \label{vis:Dual-level Quality Modulation branch (DQM)}
  \end{minipage}
\end{figure}

\textbf{Self-Conditioned Modulation (SCM):} The SCM block is designed to seamlessly integrate one-dimensional embedding and spatial priors. Specifically,  it is incorporated after each layer of the UNet's encoder and decoder. As shown in Fig.~\ref{Self-Adaptive Modulation Block (SCM)}, for the $l$-th at $t$ step, the SCM block takes the intermediate features $f_t^l$, the time embedding $t$, and the conditions $c^l$ as inputs. Recognizing that the noisy intermediate $x_t$ itself serves as a coarse mask, we concatenate a resized $x_t^l$ with $c^l$ to form the combined conditions ${\hat{c}}^l = \text{concat}(x_t^l, c^l)$, enabling self-conditioning. Subsequently, the feature map $f_t^l$ is modulated by the combined conditions ${\hat{c}}^l$ and the time embedding $t$, ultimately yielding the modulated feature $\hat{f_t^l}$:

\begin{align}
\hat{c}_{t}^l &= \text{AdaIN}({\hat{c}}^l, t),\\
\gamma, \beta  &= \text{Conv}(\hat{c}_{t}^l), \\ 
\hat{f_t^l} &= \gamma \cdot \frac{f_t^l-\mu(f_t^l)}{\sigma(f_t^l)} + \beta,
\end{align}
where the AdaIN denotes adaptive instance normalization~\cite{huang2017arbitrary}, and the Conv operation is to produce the scale $\gamma$ and shift $\beta$ with identical shape to intermediate features $f_t^l$.

\textbf{Dual-level Quality Modulation (DQM):} As shown in Fig.~\ref{vis:Dual-level Quality Modulation branch (DQM)}, the DQM is introduced to further refine the diffusion trajectory by evaluating the segmentation quality at two levels. At the error map level, the DQM takes the noisy intermediate $x_t$ and the query image $I_q$ as inputs, processing them through a convolutional block to predict the estimated error map $E_{t}$. This error map is supervised by the ground-truth error map, denoted as $E_{gt}$, which is derived by subtracting $x_t$ from the ground-truth segmentation mask, i.e., $\mathcal{L}_{em} =  \|E_{gt} - E_{t}\|_1 $, where $E_{gt} = x_0 - x_t$.

At the IoU prediction level, the DQM utilizes $x_t$, the query image $I_q$, and the predicted error map $E_{t}$ as inputs. These inputs are processed through an IoU regression branch to predict the IoU value, which is supervised by ${iou}_{gt}$. Here, ${iou}_{gt}$ is computed between the binarized versions of $x_t$ and the ground-truth mask $x_0$ using the intersection-over-union metric, and the prediction is optimized using $L_1$ loss: $\mathcal{L}_{iou} = \|{iou}_{gt} - {iou}\|_1$. where ${iou}_{gt}$ is computed between the binarized versions of $x_t$ and the ground-truth mask $x_0$, ${iou}_{gt} = ComputeIoU(x_t, x_0)$.

Subsequently, the predicted IoU value is mapped to an IoU embedding, similar to the time embedding. The resized error map $E_t^l$, modulated UNet features $\hat{f_t^l}$, and IoU embedding are then fed into a new SCM block, where $E_t^l$ serves as a condition and the IoU embedding replaces the time embedding. This process generates new quality-aware features, denoted as $\tilde{f_t^l}$, which subsequently serve as input for the $(l+1)$-th UNet layer.

\begin{align}
\hat{E}_{t}^l &= \text{AdaIN}(E_t^l, IoU),\\
\gamma, \beta &= \text{Conv}(\hat{E}_{t}^l), \\
\tilde{f_t^l} &= \gamma \cdot \frac{\hat{f_t^l} - \mu(\hat{f_t^l})}{\sigma(\hat{f_t^l})} + \beta.
\end{align}

\subsubsection{Total Loss}
The total loss is the weighted sum of the diffusion loss, the error map estimation loss, and the IoU estimation loss:
\begin{equation}
\mathcal{L}_{total} = \mathcal{L}_{diff} + \lambda_{em} \mathcal{L}_{em}  +  \lambda_{iou} \mathcal{L}_{iou},
\end{equation}
where $\lambda_{em}$ and $\lambda_{iou}$ are the weights for the error map and IoU loss, respectively.

\section{Experiments}

\subsection{Datasets and Evaluation Metrics}
\textbf{Datasets}: Following previous works~\cite{wang2019panet,tian2020prior,liu2020part,shi2022dense,min2021hypercorrelation}, we evaluated our DiffUp using two benchmarks: $\text{PASCAL-}5^i$~\cite{shaban2017one} and $\text{COCO-}20^i$~\cite{nguyen2019feature}. $\text{PASCAL-}5^i$ is built on PASCAL VOC 2012~\cite{everingham2010pascal}, enhanced by the SBD dataset~\cite{hariharan2011semantic}, and includes 20 object classes divided into 4 folds of 5 classes each. $\text{COCO-}20^i$ is derived from MSCOCO~\cite{lin2014microsoft}, comprising over 120,000 images across 80 categories, also split into 4 folds with 20 classes per fold. 

\textbf{Evaluation Metrics}: We used a leave-one-out cross-validation strategy, training on selected folds and evaluating on the remaining fold to test generalization on unseen classes. We utilized mean Intersection-over-Union ($\text{mIoU} = \frac{1}{C} \sum_{c=1}^{C} \text{IoU}_c$) and Foreground-Background IoU ($\text{FB-IoU} = \frac{1}{2} (\text{IoU}_\text{F} + \text{IoU}_\text{B})$) to assess performance.

\subsection{Implementation Details}

All experiments were implemented using the PyTorch framework on 4 NVIDIA GeForce RTX 3090 GPUs. The diffusion process utilized an input prior at resolution of $128 \times 128$, centralized to the range $[-1, 1]$. We utilized CLIP-RN50, CLIP-RN101, CLIP-ViT-B/16, IN1K-RN50, and IN1K-RN101 to evaluate the flexibility of our backbone upgrade without re-training. The trainable parameters of DiffUp was 4.2M in total. For optimization and scheduling, we trained our model using an episodic training scheme for 400 epochs on both the $\text{PASCAL-}5^i$ and $\text{COCO-}20^i$ datasets. Adam was used as the optimizer with a learning rate of $1 \times 10^{-4}$. During training, we used DDPM~\cite{ho2020denoising} with 250 steps. The training process was completed in 4.4 GPU days, with the peak GPU memory utilization reaching 20GB per GPU during training. For sampling, we applied DDIM~\cite{song2020denoising} with 50 steps, performing multiple samplings on the same inputs followed by an ensemble.

\begin{table}[t]
  \centering
   \caption{Comparison of the proposed DiffUp with the current SOTA on $\text{PASCAL-}5^i$ ~\cite{shaban2017one}. Best results are in \textcolor{red}{\textbf{red}} within each backbone. DiffUp is the first and only method that enables seamless backbone upgrade without re-training in the FSS field.}
  \label{tab:pascal_sota}
      \scalebox{0.48}{
      \begin{tabular}{clc|cccccc|cccccc|c}
        \toprule
        \multirow{2}{*}{\shortstack{\textbf{Backbone}}} & \multirow{2}{*}{\textbf{Method}} & \multirow{2}{*}{\textbf{Publication}} & \multicolumn{6}{c}{\textbf{1-shot}} & \multicolumn{6}{c|}{\textbf{5-shot}} & \multirow{2}{*}{\shortstack{\textbf{Backbone Upgrade} \\ \textbf{w/o re-train} }} \\ 

        & & & $5^{0}$ & $5^{1}$ & $5^{2}$ & $5^{3}$ & \textbf{mIoU} & \textbf{FB-IoU} & $\mathbf{5^{0}}$ & $\mathbf{5^{1}}$ & $\mathbf{5^{2}}$ & $\mathbf{5^{3}}$ & \textbf{mIoU} & \textbf{FB-IoU}  \\
        \midrule 
              
        \multirow{15}{*}{RN50}  & PFENet~\cite{tian2020prior}&  TPAMI'20  & 61.7 & 69.5 & 55.4 & 56.3 & 60.8 & 73.3 & 63.1 & 70.7 & 55.8 & 57.9 & 61.9 & 73.9 & \XSolidBrush \\
        & HSNet~\cite{min2021hypercorrelation} & ICCV'21   & 64.3 & 70.7 & 60.3 & 60.5 & 64.0 & 76.7 & 70.3 & 73.2 & 67.4 & 67.1 & 69.5 & 80.6 & \XSolidBrush \\
        & SSP~\cite{fan2022self}  & ECCV'22   & 60.5 & 67.8& 66.4 & 51.0& 61.4& - & 67.5 & 72.3 & \textcolor{red}{\textbf{75.2}} & 62.1 & 69.3 & - & \XSolidBrush\\
        & DCAMA~\cite{shi2022dense} & ECCV'22  & 67.5 & 72.3 & 59.6 & 59.0 & 64.6 & 75.7 & 70.5 & 73.9 & 63.7 & 65.8 & 68.5 & 79.5 & \XSolidBrush \\   
        & HPA\cite{cheng2023hpa} & TPAMI'23  & 65.9 & 72.0 & 64.7 & 56.8 & 64.8 & 76.4 & 70.5 & 73.3 & 68.4 & 63.4 & 68.9 & 81.1 & \XSolidBrush \\  
        & FECANet~\cite{liu2023fecanet} & TMM'23  & 69.2 & 72.3 & 62.4 & 65.7 & 67.4 & 78.7 & 72.9 & 74.0 & 65.2 & 67.8 & 70.0 & 80.7 & \XSolidBrush \\
        & ABCNet~\cite{Wang_2023_CVPR} & CVPR'23 &68.8&73.4&62.3&59.5&66.0&76.0&71.7&74.2&65.4&67.0&69.6&80.0 & \XSolidBrush \\  
        & CDPN~\cite{guo2023clip}&Entropy'23& 63.5 & 67.8 & 67.9 & 52.2 & 62.9 & - & 69.2 & 73.0 & 75.1 & 61.4 & 69.7 & - & \XSolidBrush \\
        &DCP~\cite{lang2024few} & IJCV'24&67.2&72.9&65.2&59.4&66.1&-&70.5&75.3&68.0&67.7&70.3&-& \XSolidBrush \\
        &RiFeNet~\cite{bao2024relevant} & AAAI'24&68.4&73.5&67.1&59.4&67.1&-&70.0&74.7&69.4&64.2&69.6&-& \XSolidBrush \\
        &UMTFSS~\cite{li2024label} & AAAI'24&68.3&71.3&60.0&60.7&65.1&-&71.5&74.5&61.5&68.4&68.9&-& \XSolidBrush \\
         & CLearSeg~\cite{clearseg}& MMM'24& 69.5 & 74.7 & 67.2 & 65.3 & 69.2 & - & 70.5 & 75.6 & 72.0 & 70.1 & 72.1 & - &\XSolidBrush \\
         &LSK~\cite{LSK}& PR'24& 62.1 & 69.7 & 57.8 & 57.3 & 61.7 & - & 64.3 & 71.0 & 59.5 & 58.4 & 63.3 & - &\XSolidBrush \\

        & \cellcolor[HTML]{EFEFEF} DiffUp (ours)  & \cellcolor[HTML]{EFEFEF} - &\cellcolor[HTML]{EFEFEF} \textcolor{red}{\textbf{72.2}}  &\cellcolor[HTML]{EFEFEF} \textcolor{red}{\textbf{81.7}}  &\cellcolor[HTML]{EFEFEF} \textcolor{red}{\textbf{71.2}}  &\cellcolor[HTML]{EFEFEF} \textcolor{red}{\textbf{69.2}} &\cellcolor[HTML]{EFEFEF} \textcolor{red}{\textbf{73.6}} &\cellcolor[HTML]{EFEFEF} \textcolor{red}{\textbf{83.6}} &\cellcolor[HTML]{EFEFEF} \textcolor{red}{\textbf{73.7}}  &\cellcolor[HTML]{EFEFEF} \textcolor{red}{\textbf{81.9}}  &\cellcolor[HTML]{EFEFEF} 71.7  &\cellcolor[HTML]{EFEFEF} \textcolor{red}{\textbf{70.7}}  &\cellcolor[HTML]{EFEFEF} \textcolor{red}{\textbf{74.5}}  &\cellcolor[HTML]{EFEFEF} \textcolor{red}{\textbf{84.2}} &\cellcolor[HTML]{EFEFEF} \textcolor{red}{\Checkmark} \\
        \cline{1-16} \\[-2.0ex]

         \multirow{13}{*}{RN101} & PFENet~\cite{tian2020prior} & TPAMI'20  & 60.5 & 69.4 & 54.4 & 55.9 & 60.1 & 72.9 & 62.8 & 70.4 & 54.9 & 57.6 & 61.4 & 73.5 & \XSolidBrush \\
        & HSNet~\cite{min2021hypercorrelation} & ICCV'21    & 67.3 & 72.3 & 62.0 & 63.1 & 66.2 & 77.6 & 71.8 & 74.4 & 67.0 & 68.3 & 70.4 & 80.6 & \XSolidBrush \\
        & SSP~\cite{fan2022self}  & ECCV'22   & 63.7 & 70.1& 66.7 & 55.4& 64.0& - & 70.3 & 76.3 & 77.8 & 65.5 & 72.5 & -  & \XSolidBrush \\
        & DCAMA~\cite{shi2022dense} & ECCV'22  & 65.4 & 71.4 & 63.2 & 58.3 & 64.6 & 77.6 & 70.7 & 73.7 & 66.8 & 61.9 & 68.3 & 80.8 & \XSolidBrush \\   
        & HPA~\cite{cheng2023hpa} & TPAMI'23  & 66.4 & 72.7 & 64.1 & 59.4 & 65.6 & 76.6 & 68.0 & 74.6 & 65.9 & 67.1 & 68.9 &80.4 & \XSolidBrush \\
        & ABCNet~\cite{Wang_2023_CVPR} & CVPR'23&65.3&72.9&65.0&59.3&65.6&78.5&71.4&75.0&68.2&63.1&69.4&80.8 & \XSolidBrush \\  
        &CDPN~\cite{guo2023clip}&Entropy'23& 67.8 & 71.2 & 67.7 & 57.1 & 65.9 & - & 72.8 & 76.7 & \textcolor{red}{\textbf{81.7}} & 66.7 & 74.5 & -  & \XSolidBrush \\
        &DCP~\cite{lang2024few} & IJCV'24&68.9&74.2&63.3&62.7&67.3&-&72.1&77.1&66.5&70.5&71.5&-& \XSolidBrush \\
        &RiFeNet~\cite{bao2024relevant} & AAAI'24&68.9&73.8&66.2&60.3&67.3&-&70.4&74.5&68.3&63.4&69.2&- &\XSolidBrush \\
        &LSK~\cite{LSK}& PR'24& 61.8 & 69.3 & 58.5 & 57.6 & 61.8 & - & 63.9 & 72.6 & 60.4 & 60.2 & 64.2 & - &\XSolidBrush \\

        &\cellcolor[HTML]{EFEFEF} DiffUp (ours)  &\cellcolor[HTML]{EFEFEF} - & \cellcolor[HTML]{EFEFEF} \textcolor{red}{\textbf{73.3}} &\cellcolor[HTML]{EFEFEF} \textcolor{red}{\textbf{81.3}} &\cellcolor[HTML]{EFEFEF} \textcolor{red}{\textbf{72.1}} &\cellcolor[HTML]{EFEFEF} \textcolor{red}{\textbf{73.0}} &\cellcolor[HTML]{EFEFEF} \textcolor{red}{\textbf{74.9}} &\cellcolor[HTML]{EFEFEF} \textcolor{red}{\textbf{84.8}} &\cellcolor[HTML]{EFEFEF} \textcolor{red}{\textbf{73.9}} &\cellcolor[HTML]{EFEFEF} \textcolor{red}{\textbf{81.7}} &\cellcolor[HTML]{EFEFEF} 73.2 &\cellcolor[HTML]{EFEFEF} \textcolor{red}{\textbf{73.9}}  &\cellcolor[HTML]{EFEFEF} \textcolor{red}{\textbf{75.7}} &\cellcolor[HTML]{EFEFEF} \textcolor{red}{\textbf{85.3}} &\cellcolor[HTML]{EFEFEF} \textcolor{red}{\Checkmark} \\
        \midrule 

        \multirow{4}{*}{ViT-Base} &  CLIPSeg(PC)~\cite{luddecke2022image} & CVPR'22 & - & - & - & - & 52.3 & - & - & - & - & - & - & - & \XSolidBrush \\
        &  CLIPSeg(PC+)~\cite{luddecke2022image} & CVPR'22 & - & - & - & - & 59.5 & - & - & - & - & - & - & - & \XSolidBrush \\
        &VRP-SAM~\cite{sun2024vrp}&CVPR'24& 73.9 & 78.3 & \textcolor{red}{\textbf{70.6}} & 65.0 & 71.9 & - & - & -& - & - & - & - & \XSolidBrush \\
         & \cellcolor[HTML]{EFEFEF} DiffUp (ours)  &\cellcolor[HTML]{EFEFEF}  - &\cellcolor[HTML]{EFEFEF}  \textcolor{red}{\textbf{74.8}} &\cellcolor[HTML]{EFEFEF}  \textcolor{red}{\textbf{83.0}} &\cellcolor[HTML]{EFEFEF}  70.3 &\cellcolor[HTML]{EFEFEF}  \textcolor{red}{\textbf{70.2}} &\cellcolor[HTML]{EFEFEF}  \textcolor{red}{\textbf{74.6}} &\cellcolor[HTML]{EFEFEF}  \textcolor{red}{\textbf{83.3}}  &\cellcolor[HTML]{EFEFEF}  \textcolor{red}{\textbf{76.6}}  &\cellcolor[HTML]{EFEFEF}  \textcolor{red}{\textbf{83.2}} &\cellcolor[HTML]{EFEFEF}  \textcolor{red}{\textbf{71.8}} &\cellcolor[HTML]{EFEFEF}  \textcolor{red}{\textbf{71.6}} &\cellcolor[HTML]{EFEFEF} \textcolor{red}{\textbf{75.8}}  &\cellcolor[HTML]{EFEFEF} \textcolor{red}{\textbf{84.0}} &\cellcolor[HTML]{EFEFEF}  \textcolor{red}{\Checkmark} \\
        \bottomrule

      \end{tabular}
      }

\end{table}

 \subsection{Comparison with State-of-the-Art Methods}

 \subsubsection{Pixel-wise Few-shot Segmentation Task}

Table~\ref{tab:pascal_sota} presents the mIoU and FB-IoU results on $\text{PASCAL-}5^i$ for both 1-shot and 5-shot settings. Our proposed DiffUp consistently demonstrated superior performance compared to baselines across various backbones. Specifically, with ResNet50, our DiffUp achieved an mIoU of 73.6 and 74.5 in 1-shot and 5-shot settings, respectively, surpassing the current SOTA method by a large margin. With ResNet101, DiffUp achieved an mIoU of 74.9 and 75.7 in 1-shot and 5-shot settings, respectively. With ViT-Base, DiffUp achieved an mIoU of 74.6 and 75.8 in 1-shot and 5-shot settings, respectively. Notably, DiffUp is the first and only method that enables seamless backbone upgrade without re-training in the FSS field.

\begin{table}[t]
  \centering
  \caption{Comparison of the proposed DiffUp with the SOTA on $\text{COCO-}20^i$ dataset~\cite{nguyen2019feature}. Best results are in \textcolor{red}{\textbf{red}}.}
  \label{tab:coco_sota}
      \scalebox{0.48}{
      \begin{tabular}{clc|cccccc|cccccc|c}
        \toprule
        \multirow{2}{*}{\shortstack{\textbf{Backbone}}} & \multirow{2}{*}{\textbf{Method}} & \multirow{2}{*}{\textbf{Publication}} & \multicolumn{6}{c}{\textbf{1-shot}} & \multicolumn{6}{c|}{\textbf{5-shot}}  & \multirow{2}{*}{\shortstack{ \textbf{Backbone Upgrade} \\ \textbf{w/o re-train} }}  \\ 
        & & & $\mathbf{20^{0}}$ & $\mathbf{20^{1}}$ & $\mathbf{20^{2}}$ & $\mathbf{20^{3}}$ & \textbf{mIoU} & \textbf{FB-IoU} & $\mathbf{20^{0}}$ & $\mathbf{20^{1}}$ & $\mathbf{20^{2}}$ & $\mathbf{20^{3}}$ & \textbf{mIoU} & \textbf{FB-IoU}  \\

        \midrule
        \multirow{15}{*}{RN50}& PFENet~\cite{tian2020prior} & TPAMI'20   & 36.5 & 38.6 & {34.5} & {33.8} & {35.8} & - & 36.5 & 43.3 & 37.8 & 38.4 & 39.0 & -  &\XSolidBrush  \\ 
        & HSNet~\cite{min2021hypercorrelation} & ICCV'21    & 36.3 & 43.1 & 38.7 & 38.7 & 39.2 & 68.2 & 43.3 & 51.3 & 48.2 & 45.0 & 46.9 &70.7  & \XSolidBrush \\
        & SSP~\cite{fan2022self}  & ECCV'22   & 35.5 & 39.6& 37.9 & 36.7& 37.4& - & 40.6 & 47.0 & 45.1 & 43.9 & 44.1 & -   &\XSolidBrush \\
        & DCAMA~\cite{shi2022dense} & ECCV'22  & 41.9 & 45.1 & 44.4 & 41.7 & 43.3 & 69.5 & 45.9 & 50.5 & 50.7 & 46.0 & 48.3 & 71.7  &\XSolidBrush \\   
        & HPA~\cite{cheng2023hpa} & TPAMI'23  & 40.3 & 46.6 & 44.1 & 42.7 & 43.4 & 68.2 & 45.5 & 55.4 & 48.9 & 50.2 & 50.0 & 71.2  &\XSolidBrush  \\  
        & FECANet~\cite{liu2023fecanet} & TMM'23  & 38.5 & 44.6 & 42.6 & 40.7 & 41.6 & 69.6 & 44.6 & 51.5 & 48.4 & 45.8 & 47.6 & 71.1  &\XSolidBrush  \\   
        &ABCNet~\cite{Wang_2023_CVPR} & CVPR'23&42.3&46.2&46.0&42.0&44.1&69.9&45.5&51.7&52.6&46.4&49.1&72.7 &\XSolidBrush\\    
        & CDPN~\cite{guo2023clip}&Entropy'23& 48.3 & 36.5 & 28.9 & 26.5 & 35.0 & - & \textcolor{red}{\textbf{56.5}} & 43.9 & 38.0 & 35.6 & 43.5 & - & \XSolidBrush \\       
        &AdaptiveFSS~\cite{wang2024adaptive} & AAAI'24&44.1&55.0&46.5&48.5&48.5&-&48.1&60.8&54.8&51.9&53.9&- &\XSolidBrush\\
        &RiFeNet~\cite{bao2024relevant} & AAAI'24&39.1&47.2&44.6&45.4&44.1&-&44.3&52.4&49.3&48.4&48.6&- &\XSolidBrush\\
        & CLearSeg~\cite{clearseg}& MMM'24&43.7&55.7&48.0&48.0&48.9&-&47.3&\textcolor{red}{\textbf{61.8}}&54.3&53.8&54.3&- &\XSolidBrush\\
        &LSK~\cite{LSK}& PR'24& 35.3 & 39.0 & 38.7 & 33.4 & 36.6 & - & 37.4 & 41.2 & 39.1 & 34.2 & 38.0 & - &\XSolidBrush \\       
        & \cellcolor[HTML]{EFEFEF} DiffUp (ours)  & \cellcolor[HTML]{EFEFEF} -&\cellcolor[HTML]{EFEFEF} \textcolor{red}{\textbf{46.7}} &\cellcolor[HTML]{EFEFEF} \textcolor{red}{\textbf{56.8}} &\cellcolor[HTML]{EFEFEF} \textcolor{red}{\textbf{58.5}} &\cellcolor[HTML]{EFEFEF} \textcolor{red}{\textbf{55.8}} &\cellcolor[HTML]{EFEFEF} \textcolor{red}{\textbf{54.5}} &\cellcolor[HTML]{EFEFEF} \textcolor{red}{\textbf{76.7}} &\cellcolor[HTML]{EFEFEF} 52.4 &\cellcolor[HTML]{EFEFEF} 58.6 &\cellcolor[HTML]{EFEFEF} \textcolor{red}{\textbf{59.3}} &\cellcolor[HTML]{EFEFEF} \textcolor{red}{\textbf{56.9}} &\cellcolor[HTML]{EFEFEF} \textcolor{red}{\textbf{56.8}} &\cellcolor[HTML]{EFEFEF} \textcolor{red}{\textbf{77.9}} &\cellcolor[HTML]{EFEFEF} \textcolor{red}{\Checkmark} \\
        \cline{1-16} \\[-2.0ex]

        \multirow{9}{*}{RN101} & PFENet~\cite{tian2020prior} & TPAMI'20  & {36.8} & {41.8} & {38.7} & {36.7} & {38.5} & {63.0} & {40.4} & {46.8} & {43.2} & {40.5} & {42.7} & {65.8}  &\XSolidBrush \\
        & HSNet~\cite{min2021hypercorrelation} & ICCV'21   & {37.2} & {44.1} & {42.4} & {41.3} & {41.2} & {69.1} & {45.9} & {53.0} & {51.8} & {47.1} & {49.5} & {72.4}  &\XSolidBrush\\
        & SSP~\cite{fan2022self}  & ECCV'22   & 39.1 & 45.1& 42.7 & 41.2& 42.0& - & 47.4 & 54.5 & 50.4 & 49.6 & 50.2 & -  &\XSolidBrush \\
        & DCAMA~\cite{shi2022dense} & ECCV'22  & 41.5 & 46.2 & 45.2 & 41.3 & 43.5 & 69.9 & 48.0 & 58.0 & 54.3 & 47.1 & 51.9 & 73.3  &\XSolidBrush \\  
        & HPA\cite{cheng2023hpa} & TPAMI'23  & 43.1 & 50.0 & 44.8 & 45.2 & 45.8 & 68.4 & 49.2 & 57.8 & 52.0 & 50.6 & 52.4 & 74.0  &\XSolidBrush \\  
        &CDPN~\cite{guo2023clip}&Entropy'23& 52.3 & 40.7 & 33.7 & 31.7 & 39.6 & - & \textcolor{red}{\textbf{61.5}} & 48.4 & 42.7 & 43.4 & 49.0 & -  & \XSolidBrush \\
        &LSK~\cite{LSK}& PR'24& 37.6 & 40.2 & 38.8 & 36.1 & 38.2 & - & 40.5 & 41.6 & 42.3 & 37.3 & 40.4 & - &\XSolidBrush \\
        &\cellcolor[HTML]{EFEFEF} DiffUp (ours)  &\cellcolor[HTML]{EFEFEF} - & \cellcolor[HTML]{EFEFEF} \textcolor{red}{\textbf{50.6}} &\cellcolor[HTML]{EFEFEF} \textcolor{red}{\textbf{59.8}} &\cellcolor[HTML]{EFEFEF} \textcolor{red}{\textbf{55.9}} &\cellcolor[HTML]{EFEFEF} \textcolor{red}{\textbf{59.2}} &\cellcolor[HTML]{EFEFEF} \textcolor{red}{\textbf{56.4}} &\cellcolor[HTML]{EFEFEF} \textcolor{red}{\textbf{77.4}} &\cellcolor[HTML]{EFEFEF} 53.4 &\cellcolor[HTML]{EFEFEF} \textcolor{red}{\textbf{62.7}} &\cellcolor[HTML]{EFEFEF} \textcolor{red}{\textbf{59.8}} &\cellcolor[HTML]{EFEFEF} \textcolor{red}{\textbf{61.2}}  &\cellcolor[HTML]{EFEFEF} \textcolor{red}{\textbf{59.3}} &\cellcolor[HTML]{EFEFEF} \textcolor{red}{\textbf{79.1}} &\cellcolor[HTML]{EFEFEF} \textcolor{red}{\Checkmark} \\

        \cline{1-16} \\[-2.0ex]
        \multirow{4}{*}{ViT-Base} &  CLIPSeg(PC)~\cite{luddecke2022image} & CVPR'22 & - & - & - & - & 33.2 & - & - & - & - & - & - & - & \XSolidBrush \\
        &  CLIPSeg(PC+)~\cite{luddecke2022image} & CVPR'22 & - & - & - & - & 33.3 & - & - & - & - & - & - & - & \XSolidBrush \\
        &VRP-SAM~\cite{sun2024vrp}&CVPR'24& 48.1 & 55.8 & \textcolor{red}{\textbf{60.6}} & 51.6 & 53.9 & - & - &- & - & - & - & - & \XSolidBrush \\
        & \cellcolor[HTML]{EFEFEF} DiffUp (ours)  &\cellcolor[HTML]{EFEFEF}  - &\cellcolor[HTML]{EFEFEF}  \textcolor{red}{\textbf{48.6}} &\cellcolor[HTML]{EFEFEF}  \textcolor{red}{\textbf{59.4}} &\cellcolor[HTML]{EFEFEF}  54.3 &\cellcolor[HTML]{EFEFEF}  \textcolor{red}{\textbf{58.4}} &\cellcolor[HTML]{EFEFEF}  \textcolor{red}{\textbf{55.2}} &\cellcolor[HTML]{EFEFEF}  \textcolor{red}{\textbf{77.3}} &\cellcolor[HTML]{EFEFEF}  \textcolor{red}{\textbf{51.6}} &\cellcolor[HTML]{EFEFEF}  \textcolor{red}{\textbf{63.2}} &\cellcolor[HTML]{EFEFEF}  \textcolor{red}{\textbf{57.5}} &\cellcolor[HTML]{EFEFEF}  \textcolor{red}{\textbf{59.1}} &\cellcolor[HTML]{EFEFEF}  \textcolor{red}{\textbf{57.9}} &\cellcolor[HTML]{EFEFEF}  \textcolor{red}{\textbf{78.2}} &\cellcolor[HTML]{EFEFEF}  \textcolor{red}{\Checkmark} \\

        \bottomrule
      \end{tabular}
      }

\end{table}

Table~\ref{tab:coco_sota} demonstrates significant and consistent improvements across the $\text{COCO-}20^i$ dataset~\cite{nguyen2019feature}. Using the ResNet50 backbone, our DiffUp approach achieved state-of-the-art performance with mIoU scores of 54.5 and 56.8 for 1-shot and 5-shot settings, respectively. When implemented with the ResNet101 backbone, DiffUp further improved to reach mIoU scores of 56.4 and 59.3 for 1-shot and 5-shot settings, respectively. These results substantially outperform previous SOTA by a considerable margin.

\subsubsection{Weakly-supervised Few-shot Segmentation Task}

\begin{table}[t]
  \centering
  \caption{The DiffUp effectively handles diverse weak annotations, including bounding box, scribble, FSS without support mask, and purely textual annotation. Moreover, it excels in cross-dataset setting. $^{*}$ indicates testing via ourselves.}
  \label{tab:bbox-scribble-coseg-textual-crossdomain} 
      \scalebox{0.8}{
      \begin{tabular}{cl|cccccc}
              \toprule
              \textbf{Task} & {\textbf{Method}} &  $\mathbf{5^{0}}$ & $\mathbf{5^{1}}$ & $\mathbf{5^{2}}$ & $\mathbf{5^{3}}$ & \textbf{mIoU}  & \textbf{FB-IoU} \\

              \midrule
              \multirow{3}{*}{\shortstack{bbox-level\\FSS task}}
              & HSNet~\cite{min2021hypercorrelation}$^{*}$   & 53.4 & 64.5 & 52.7 & 51.8 & 55.6 & 70.1  \\
              & DCAMA~\cite{shi2022dense} $^{*}$  & 62.2 & 70.3 & 56.3 & 56.0 & 61.2 & 73.0  \\
              & \cellcolor[HTML]{EFEFEF} DiffUp (ours) &\cellcolor[HTML]{EFEFEF}  \textcolor{red}{\textbf{70.7}}  &\cellcolor[HTML]{EFEFEF} \textcolor{red}{\textbf{81.1}}  & \cellcolor[HTML]{EFEFEF} \textcolor{red}{\textbf{69.6}}  & \cellcolor[HTML]{EFEFEF}  \textcolor{red}{\textbf{67.7}}  & \cellcolor[HTML]{EFEFEF} \textcolor{red}{\textbf{72.3}}  & \cellcolor[HTML]{EFEFEF}\textcolor{red}{\textbf{82.5}} \\

                \midrule
              \multirow{3}{*}{\shortstack{scribble-level\\FSS task}}
              & HSNet~\cite{min2021hypercorrelation}$^{*}$   & 47.6  & 64.2  & 50.5  & 54.3  & 54.1  & 70.8   \\
              & DCAMA~\cite{shi2022dense} $^{*}$  &  58.3  & 69.8  & 53.6  & 57.3  & 59.8  & 73.7 \\

              \cline{2-8} \\[-2.0ex]
              &   \cellcolor[HTML]{EFEFEF} DiffUp (ours)   &  \cellcolor[HTML]{EFEFEF} \textcolor{red}{\textbf{67.1 }} & \cellcolor[HTML]{EFEFEF}  \textcolor{red}{\textbf{79.1}} & \cellcolor[HTML]{EFEFEF} \textcolor{red}{\textbf{67.6}} & \cellcolor[HTML]{EFEFEF} \textcolor{red}{\textbf{67.3}} & \cellcolor[HTML]{EFEFEF} \textcolor{red}{\textbf{70.3}} &  \cellcolor[HTML]{EFEFEF} \textcolor{red}{\textbf{81.5}}  \\

              \midrule
              \multirow{5}{*}{\shortstack{FSS task \\ w/o \\support mask}}
              & (V+S)-1~\cite{siam2020weakly} & 49.5 & 65.5 & 50.0 & 49.2 & 53.5 & 65.6  \\
              & (V+S)-2~\cite{siam2020weakly} & 42.5 & 64.8 & 48.1 & 46.5 & 50.5 & 64.1  \\
              & IMR-HSNet~\cite{wang2023iterative}   & {62.6} & {69.1} & {56.1} & {56.7} & {61.1} & -  \\
              & HSNet~\cite{min2021hypercorrelation}-RN101 &  \textcolor{red}{\textbf{66.2}} & {69.5} & {53.9} & {56.2} & {61.5} & 72.5  \\
              \cline{2-8} \\[-2.0ex]
              &  \cellcolor[HTML]{EFEFEF}DiffUp (ours)   &\cellcolor[HTML]{EFEFEF} 62.2 &\cellcolor[HTML]{EFEFEF}  \textcolor{red}{\textbf{73.7}} &\cellcolor[HTML]{EFEFEF} \textcolor{red}{\textbf{63.2}} &\cellcolor[HTML]{EFEFEF}  \textcolor{red}{\textbf{57.2}} &\cellcolor[HTML]{EFEFEF}  \textcolor{red}{\textbf{64.1}} &\cellcolor[HTML]{EFEFEF}  \textcolor{red}{\textbf{77.2}}   \\

                \midrule
                \multirow{4}{*}{\shortstack{textual\\annotation }}

                &LSeg~\cite{li2022languagedriven}-RN101 & 52.8&53.8&44.4&38.5&47.4&64.1  \\ 
                &LSeg~\cite{li2022languagedriven}-ViT-L &\textcolor{red}{\textbf{61.3}}&63.6&43.1&41.0&52.3&67.0 \\ 
                &SAZS~\cite{liu2023delving}-DRN &57.3&60.3&\textcolor{red}{\textbf{58.4}}&45.9&55.5&66.4 \\ 
                \cline{2-8} \\[-2.0ex]
                &\cellcolor[HTML]{EFEFEF} DiffUp (ours) &\cellcolor[HTML]{EFEFEF} 52.6 &\cellcolor[HTML]{EFEFEF} \textcolor{red}{\textbf{71.6}} &\cellcolor[HTML]{EFEFEF} 54.0 &\cellcolor[HTML]{EFEFEF}  \textcolor{red}{\textbf{53.8}} &\cellcolor[HTML]{EFEFEF}  \textcolor{red}{\textbf{58.0}} &\cellcolor[HTML]{EFEFEF}  \textcolor{red}{\textbf{67.7}}  \\

              \bottomrule
      \end{tabular}
      }

\end{table}

Table~\ref{tab:bbox-scribble-coseg-textual-crossdomain} showcases the effectiveness of DiffUp under various weak annotations, including bounding box, scribble, support image without corresponding mask, and text. These weak supervisions were unified into a common form of spatial priors, thus enabling inference within a single network. For \textbf{\textit{bbox-level}} annotation, DiffUp achieved an mIoU of 72.3, surpassing both HSNet~\cite{min2021hypercorrelation} and DCAMA~\cite{shi2022dense}. In the \textbf{\textit{scribble-level}} FSS task, DiffUp reached an mIoU of 70.3. When only support image \textbf{\textit{without corresponding support mask}} was available, a scenario similar to co-segmentation task, DiffUp attained an mIoU of 64.1. Furthermore, in the absence of any visual cues and relying solely on \textbf{\textit{textual descriptions}} to indicate the segmentation region, DiffUp achieved an mIoU of 58.0, surpassing the 55.5 mIoU attained by the previous text-based ZSS methods~\cite{liu2023delving,li2022languagedriven}. Experiments with these weak annotations demonstrated the effectiveness of DiffUp in treating weak annotations with different granularities as noisy intermediates in a diffusion process.

\subsubsection{Cross-dataset Few-shot Segmentation Task}

\begin{table}[htbp]
  \centering
  \caption{Comparison on cross-dataset setting (COCO-20$^i$ to PASCAL-5$^i$).}
  \label{tab:crossdomain} 
      \scalebox{0.9}{
      \begin{tabular}{cl|cccccc}
              \toprule
              \textbf{Task} & {\textbf{Method}} &  $\mathbf{5^{0}}$ & $\mathbf{5^{1}}$ & $\mathbf{5^{2}}$ & $\mathbf{5^{3}}$ & \textbf{mIoU} & \textbf{$\triangle$} \\

              \multirow{4}{*}{\shortstack{cross-dataset\\FSS task}}
              & HSNet~\cite{min2021hypercorrelation}   & 48.7 & 61.5 & 63.0 & 72.8 & 61.5 & - \\
              & CWT~\cite{lu2021simpler}    & 53.5 & 59.2 & 60.2 & 64.9 & 59.4 & -2.1 \\
              & CDFSS~\cite{wang2022remember}   & \textcolor{red}{\textbf{57.4}} & 62.2 & 68.0 & 74.8 & 65.6 & +4.1 \\
              \cline{2-8} \\[-2.0ex]
              &  \cellcolor[HTML]{EFEFEF} DiffUp (ours)   &\cellcolor[HTML]{EFEFEF} 51.1 &\cellcolor[HTML]{EFEFEF}  \textcolor{red}{\textbf{74.6}} &\cellcolor[HTML]{EFEFEF}  \textcolor{red}{\textbf{71.2}} &\cellcolor[HTML]{EFEFEF}  \textcolor{red}{\textbf{83.4}} &\cellcolor[HTML]{EFEFEF}  \textcolor{red}{\textbf{70.1}} &\cellcolor[HTML]{EFEFEF}  \textcolor{red}{\textbf{+8.6}} \\

              \bottomrule
      \end{tabular}
      }
\end{table}

We evaluated the cross-dataset performance on the $\text{COCO-}20^i$~\cite{nguyen2019feature} to $\text{PASCAL-}5^i$~\cite{shaban2017one} setting.  While cross-dataset generalization ability was not the core focus of DiffUp, it exhibited remarkable capabilities in this field. Quantitatively, Table~\ref{tab:crossdomain} shows a notable improvement of approximately 4.5 mIoU in 1-shot scenario, with DiffUp achieving 70.1 compared to CDFSS's 65.6~\cite{wang2022remember}.  We ascribed this robustness to DiffUp's utilization of diverse priors with minimal inter-dataset variance. This characteristic equipped the model with a domain-agnostic property,  enabling it to generalize more effectively to unseen classes.

  \subsection{Backbone Upgrade without Re-Training}

  To demonstrate DiffUp's flexibility in upgrading backbones without re-training, we conducted cross-backbone stitching using CLIP-RN50, CLIP-RN101, CLIP-ViT-B/16, and IN1K-RN50. We trained decoders on fold-3 of the $\text{PASCAL-}5^i$ dataset, then stitched them with other backbones, reporting mIoU without re-training. Table~\ref{tab:backbone_upgrade} shows DiffUp's performance gains across three backbone upgrade types: scale-up, architectural shift, and pre-training adjustments. For the backbone scale-up upgrade, the mIoU of CLIP-RN50 with its correspondingly trained decoder was 69.2. When upgrading the backbone from CLIP-RN50 to CLIP-RN101 while directly stitching the decoder, we achieved 71.9. Shifting from CNN to ViT architecture yielded 0.6 gain (69.8 mIoU). Changing from IN-1K to CLIP pre-training increased mIoU from 59.1 to 68.4. Such flexibility to upgrade and gaining improvements without re-training stemmed from DiffUp's use of backbone-agnostic multi-granularity priors in the diffusion process, a capability unprecedented in FSS methods.
  
  \begin{table}[htbp]
    \centering
    \caption{The flexibility of DiffUp in backbone upgrade without re-training on $\text{PASCAL-}5^i$, including scale-up,  architectural shift, and pre-training adjustment.}
    \label{tab:backbone_upgrade}
    \scalebox{0.8}{ 
    \begin{tabular}{cl|ccc}
      \toprule
      \multirow{2}{*}{\textbf{Types}} &    \multirow{2}{*}{\textbf{Backbone upgrade}}  & \textbf{Before} & \textbf{After} & \multirow{2}{*}{\textbf{ $\Delta$ }} \\ 
      &  & \textbf{upgrade}  & \textbf{upgrade}  &\\
      \midrule
      scale-up&CLIP-RN50 $\rightarrow$ CLIP-RN101   & 69.2 & 71.9 & +2.7 \\
      architectural shift&CLIP-RN50 $\rightarrow$ CLIP-ViT-B/16   &69.2   & 69.8 & +0.6 \\
      pre-training adjustment& IN1K-RN50 $\rightarrow$ CLIP-RN50   & 59.1  & 68.4 & +9.3 \\
      \bottomrule
    \end{tabular}
    }
  \end{table}
  
  We further investigated the backbone downgrading scenario, which is crucial for edge device deployment. Our baseline experiments established that IN1K-RN50, IN1K-RN101, and IN1K-RN152 achieved 59.1, 62.3, and 64.8 mIoU respectively. When downgrading from IN1K-RN152 to IN1K-RN101, DiffUp maintained 62.6 mIoU (96.6\% of the original performance). Conversely, upgrading from IN1K-RN50 to IN1K-RN101 yielded 61.7 mIoU, demonstrating a significant improvement of 2.6 mIoU. This demonstrates DiffUp's flexibility in both upgrading and downgrading scenarios, making it particularly suitable for resource-constrained applications.

  \begin{figure}[htbp]
    \centering
    \includegraphics[width= 0.9\linewidth]{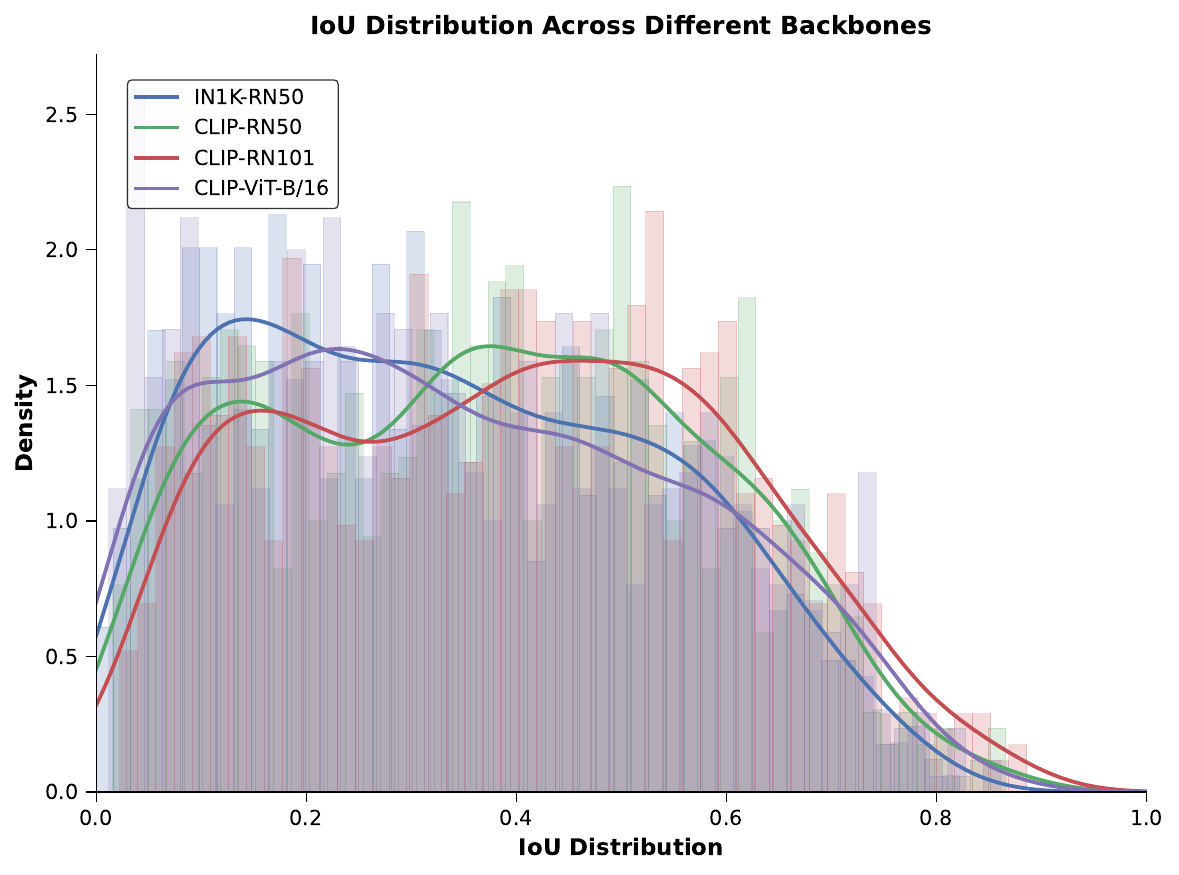}
    \caption{IoU distribution analysis of binarized probability priors across different backbones, demonstrating DiffUp's ability to standardize features from diverse backbones.}
    \label{fig:diffup_iou_distribution_4model}
  \end{figure}
  Figure~\ref{fig:diffup_iou_distribution_4model} reveals DiffUp's backbone upgrade capability through histogram analysis and kernel density estimation. Coarse probability priors from different backbones were binarized using a 0.5 threshold and IoU values were computed against ground truth masks to create statistical distributions. Despite the inherent distributional variations among CLIP-RN50, CLIP-RN101, CLIP-ViT-B/16, and IN1K-RN50 backbones, DiffUp's transformation module effectively standardized these features into a normalized [0,1] probability space, significantly reducing bias. The diffusion decoder further mitigated the residual differences through its iterative mechanism, maintaining performance stability across backbone switches.

  \subsection{Visualization of DiffUp's Segmentation Process}

  \begin{figure}[t]
    \centering
    \includegraphics[width= 1.\linewidth]{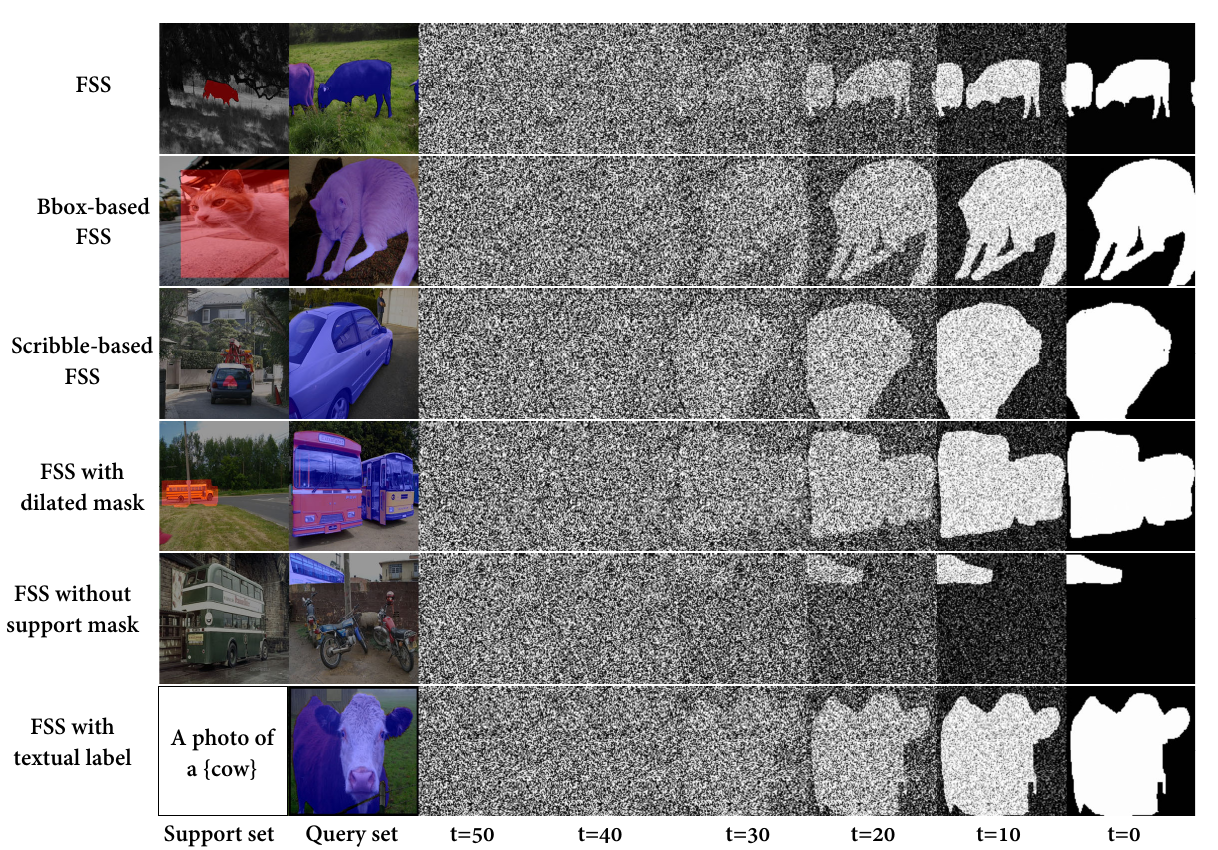}
    \caption{Visualization of segmentation results under different types of support annotations and sampling timesteps. The right side of each row illustrates the noisy intermediates at various sampling timesteps in the diffusion process. }
    \label{vis:timesteps-vis}
  \end{figure}

  To present a more detailed view of DiffUp's effectiveness, Fig.~\ref{vis:timesteps-vis} illustrates segmentation results across various support annotation types and sampling timesteps. Each row displays intermediate noise results at different timesteps during the diffusion process, progressing from right to left. The visualization covers six distinct scenarios: mask-based few-shot segmentation, bbox-based few-shot segmentation, scribble-based few-shot segmentation, few-shot segmentation with inaccurate support masks, few-shot segmentation without support masks, and text-only segmentation tasks. The columns represent segmentation results at sampling timesteps of 50 (pure noise), 40, 30, 20, 10, and 0, respectively. These visualizations demonstrate DiffUp's segmentation performance across different support annotation types and sampling timesteps, highlighting the model's versatility and robustness.

  \subsection{Ablation Studies}

  To investigate the impact of key components in our DiffUp, we conducted comprehensive ablation studies. For all ablations, we employed the CLIP-RN50 backbone on fold-3 of the $\text{PASCAL-}5^i$ dataset~\cite{shaban2017one} in 1-shot scenario.

  \begin{table}[htbp]
    \centering
    \caption{Ablation study on the affinity-enhanced strategy ($A_{ff}$) , additional backbone-agnostic query information (edge), error map modulate in DQM module ($E_t$ modulate), and IoU modulate in DQM module (IoU modulate).}
    \label{tab:ablation_aqq_edge_errormap_iou}
    \scalebox{0.9}{ 
    \begin{tabular}{ccccc|cc}
    \bottomrule
      \multirow{2}{*}{\textbf{NO.}} & \multirow{2}{*}{$A_{ff}$} & \multirow{2}{*}{\textbf{edge}} & \multicolumn{2}{c|}{\textbf{DQM}}  & 
      \multicolumn{2}{c}{\textbf{1-shot}}\\
      
      &&& $E_t$ modulate&IoU modulate & mIoU & $\nabla$ \\
      
      \midrule
      1&$\checkmark$&$\checkmark$&$\checkmark$&$\checkmark$&69.2&-\\
      2&-&$\checkmark$&$\checkmark$&$\checkmark$&68.4&-0.8\\
      3&$\checkmark$&-&$\checkmark$&$\checkmark$&67.8&-1.4\\
      4&$\checkmark$&$\checkmark$&$\checkmark$&-&68.6&-0.6\\
      5&$\checkmark$&$\checkmark$&-&$\checkmark$&67.5&-1.7\\
      6&$\checkmark$&$\checkmark$&-&-&67.2&-2.0\\
      \bottomrule
    \end{tabular}
    }
  \end{table}

  \subsubsection{Dual-level Quality Modulation (DQM)}
  To align multi-granularity priors with noisy intermediates during the diffusion process, DiffUp incorporated two levels of quality modulation: error map level and IoU level. As shown in  Table~\ref{tab:ablation_aqq_edge_errormap_iou}, without IoU-level modulation, the performance decreased to 68.6 mIoU. In the absence of any quality modulation, performance was limited to 67.2 mIoU, underscoring the effectiveness of both levels of quality modulation.

  \subsubsection{Affinity-Enhanced Strategy}
   The influence of the affinity-enhanced strategy in Equation~\ref{eq:aff_prior} was investigated through Table~\ref{tab:ablation_aqq_edge_errormap_iou}, showing an mIoU drop from $69.2$ to $68.4$, demonstrating its effectiveness in improving segmentation accuracy.

  \subsubsection{Additional Backbone-Agnostic Query Information} Our study employed additional backbone-agnostic query information, specifically edge operators, alongside transformed priors. As shown in Table~\ref{tab:ablation_aqq_edge_errormap_iou}, the absence of these additional features resulted in an mIoU decrease from 69.2 to 67.8, highlighting their significance.

  \subsection{More Discussions}

  \subsubsection{Transformation Functions}

  To investigate alternative transformation function $\Phi$ for generating coarse priors, we explored $L_1$ and $L_2$ distances with different conversions ($\exp(-d)$, $1/(1 + d)$) as replacements for the default cosine similarity. As shown in Table~\ref{transformation_functions}, cosine similarity achieves optimal performance (69.2 mIoU), equivalent to normalized $L_2$ distance with conversion $1 - 0.5 L_2^2$.
  \begin{table}[htbp]
    \centering
    \caption{Comparison on different transformation functions.} \label{transformation_functions}
    \scalebox{1}{ 
    \begin{tabular}{ccc}
    \toprule
    \textbf{Distance Metric (d)} & \textbf{Conversion Function} & \textbf{mIoU} \\
    \midrule
     Cosine (default)             & $-$               & 69.2 \\
    $\boldsymbol{L_1}$            & $\exp(-d)$        & 60.6 \\
    $\boldsymbol{L_1}$            & $1/(1+d)$         & 68.4 \\
    $\boldsymbol{L_2}$            & $1-0.5d^2$        &  69.2 \\
    $\boldsymbol{L_2}$            & $\exp(-d)$        & 69.2 \\
    $\boldsymbol{L_2}$            & $1/(1+d)$         & 69.1 \\

    multi-metric                  &multi-conversion   & \textbf{69.5} \\
    \bottomrule
    \end{tabular}}
    \end{table}
  
Furthermore, considering that different transformation functions essentially generate priors of varying granularities in the probability space, which aligns perfectly with our Unified Quality-aware Diffusion-based Decoder, we have explored a multi-metric training strategy that stochastically integrates different distance metrics ($L_1$, $L_2$, Cosine) and conversion functions ($\exp(-d)$, $1/(1 + d)$) during training. With this strategy, our DiffUp achieves 69.5 mIoU.

  \subsubsection{Architectural Choice}
  
  \begin{table}[tbp]
  \caption{Performance comparison between GAN-based and diffusion-based implementations on PASCAL-5$^i$ fold-3. Our diffusion-based approach shows consistent advantages across different annotation types.}
  \centering
  \scalebox{0.95}{
  \begin{tabular}{l|ccc|c}
  \toprule
  Method & Scribble & Box  & Mask & Average \\
  \midrule
  StyleGAN2-based & 61.6 & 62.9 & 64.7 & 63.1 \\
  DiffUp (Ours) & 67.3 & 67.7 & 69.2 &  68.1 \\
  \bottomrule
  \end{tabular}
  }
  \label{tab:gan_comparison}
  \end{table}
  
To systematically evaluate different generative frameworks (diffusion v.s. GAN) for FSS, we developed a StyleGAN2~\cite{karras2020analyzing} based implementation that mirrors our conditioning mechanism. Specifically, the architecture consisted of a prior encoding network that maps coarse priors to latent vectors, which are then concatenated with Gaussian noise for conditioning. The generator follows StyleGAN2's architecture with the final layer modified for mask prediction. As shown in Table~\ref{tab:gan_comparison}, our diffusion-based approach demonstrates consistent performance advantages across all annotation types.  This indicates that the progressive denoising process in diffusion models provides a more effective framework for handling uncertainty in coarse priors.

  \subsubsection{Sensitivity Analysis} 

  \begin{figure}[htbp]
    \centering
    \includegraphics[width= 0.85\linewidth]{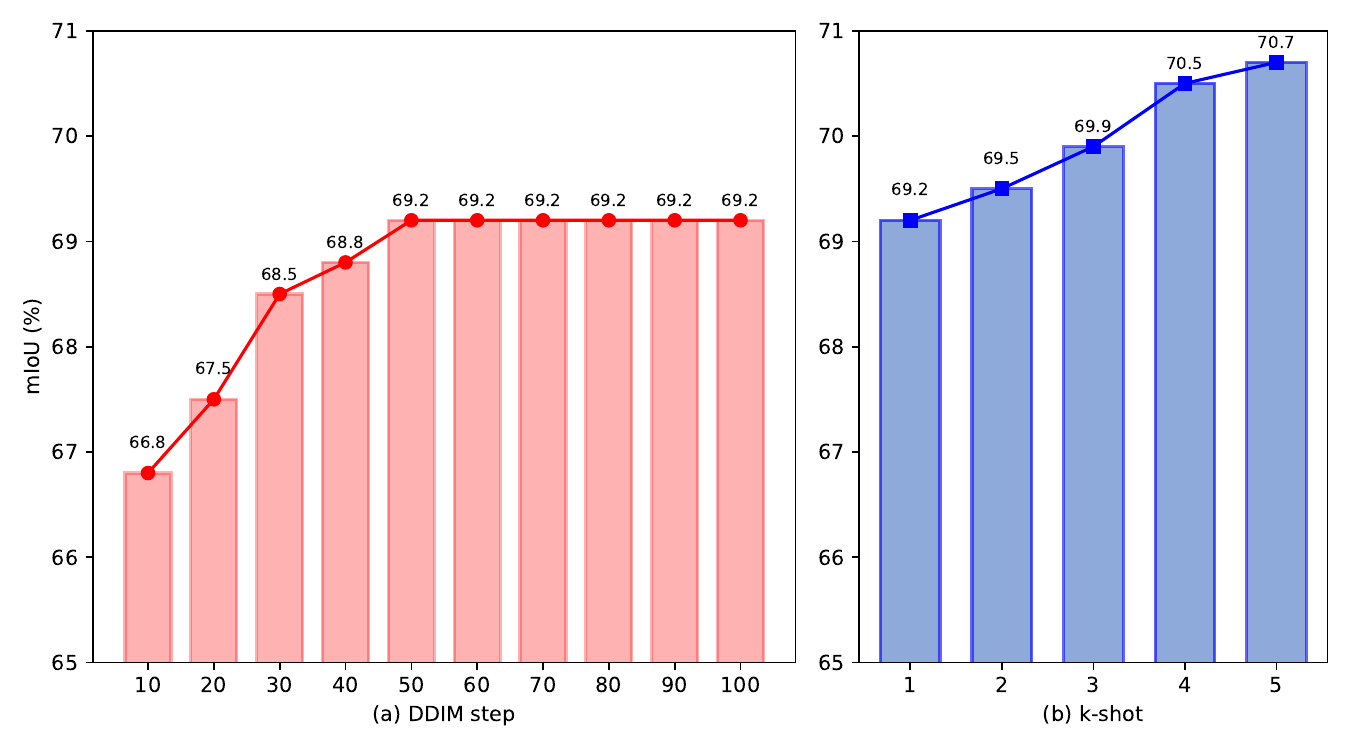}
    \caption{Sensitivity Analysis on DDIM steps and k-shot numbers.}
    \label{vis:ddim-steps-nshot}
  \end{figure}

  Furthermore, we conducted a sensitivity analysis on two key parameters: DDIM sampling steps and the k-shot number. It is worth noting that even in many-shot scenarios, our DiffUp requires only a single inference pass. As illustrated in Fig.~\ref{vis:ddim-steps-nshot}, both computational resources (reflected by sampling steps) and data availability (k-shot number) slightly influenced DiffUp's performance.

  \subsubsection{Trade-off between Backbone Flexibility and Feature Fidelity}
  Our adoption of cosine similarity as the feature transformation, while enabling backbone upgrade without re-training, introduces an inherent trade-off between feature fidelity and backbone flexibility. To quantify this trade-off, we implemented a Low-level Feature Enhancement (LFE) module that incorporates backbone-specific features into our diffusion-based decoder. The LFE module extracted fine-grained visual information from the initial blocks of RN50, and fused them with corresponding UNet features via a simple convolution block. 
  
  Experimental results in Table~\ref{tab:LFE} showed that DiffUp+LFE achieved 70.4 and 71.5 mIoU under one-shot and five-shot settings, respectively. However, this enhancement necessarily compromised the backbone-agnostic property of our framework. This empirical analysis reveals the fundamental challenge in balancing feature richness and backbone flexibility, suggesting an important direction for future research.

  \begin{table}[tbp]
    \centering
    \caption{Trade-off between backbone flexibility and feature fidelity.}\label{tab:LFE}
    \scalebox{0.8}{
    \begin{tabular}{c|cc|c}
      \toprule
      \multirow{2}{*}{\textbf{Method}}&  \multicolumn{2}{c|}{\textbf{mIoU}} & \textbf{Backbone Upgrade}\\
      &\textbf{1-shot}&\textbf{5-shot}&\textbf{w/o re-train}\\
      \midrule 
      DiffUp & 69.2 & 70.7 &\textcolor{red}{\Checkmark} \\
      DiffUp+LFE &  70.4 & 71.5 & \XSolidBrush \\
      \bottomrule
   \end{tabular}}
  \end{table}

  \subsubsection{Complexity Analysis}

  \begin{table}[htbp]
    \centering
    \caption{Comparison on parameters.}
    \label{tab:complexity}
    \scalebox{0.75}{ 
    \begin{tabular}{c|ccccc|c}
      \toprule
      \multirow{2}{*}{\textbf{Method}}&\multicolumn{5}{c|}{\textbf{learnable params (M)}} & \textbf{backbone upgrade}\\
      &\textbf{Vgg16} & \textbf{Resnet50} &\textbf{ResNet101} &\textbf{Swin-base} &\textbf{total} &\textbf{without re-training}\\
  
      \midrule
      HSNet~\cite{min2021hypercorrelation}&2.6&2.6&2.6&2.6&10.4&\XSolidBrush\\
      DCAMA~\cite{shi2022dense}&-&14.2&14.2&5.1&33.5&\XSolidBrush\\
      DiffUp(Ours)&4.2&4.2&4.2&4.2&\textcolor{red}{\textbf{4.2}}&\textcolor{red}{\Checkmark}\\
      \bottomrule
    \end{tabular}
    }
  \end{table}

  Table~\ref{tab:complexity} presents a comparative analysis of model parameters across different backbones. Previous methods required training separate decoders for each backbone, leading to significant parameter overhead when supporting multiple backbones. For instance, when deploying with Vgg16, ResNet50, ResNet101, and Swin-base backbones, HSNet~\cite{min2021hypercorrelation} requires maintaining 10.4M total learnable parameters (2.6M per backbone), while DCAMA~\cite{shi2022dense} demands an even larger parameter budget of 33.5M across its compatible architectures. In contrast, our proposed DiffUp maintains a consistent parameter count of only 4.2M, regardless of the backbone architecture. This efficiency stems from the proposed BAFT module and diffusion-based decoder design, which uniquely enables backbone upgrades without re-training, a key technical contribution absent in previous approaches. As a result, DiffUp significantly reduces the model maintenance burden while offering flexible deployment across varied network architectures.

  \subsubsection{Error Map Visualization}
  
  \begin{figure}[htbp]
    \centering
    \includegraphics[width= 1. \linewidth]{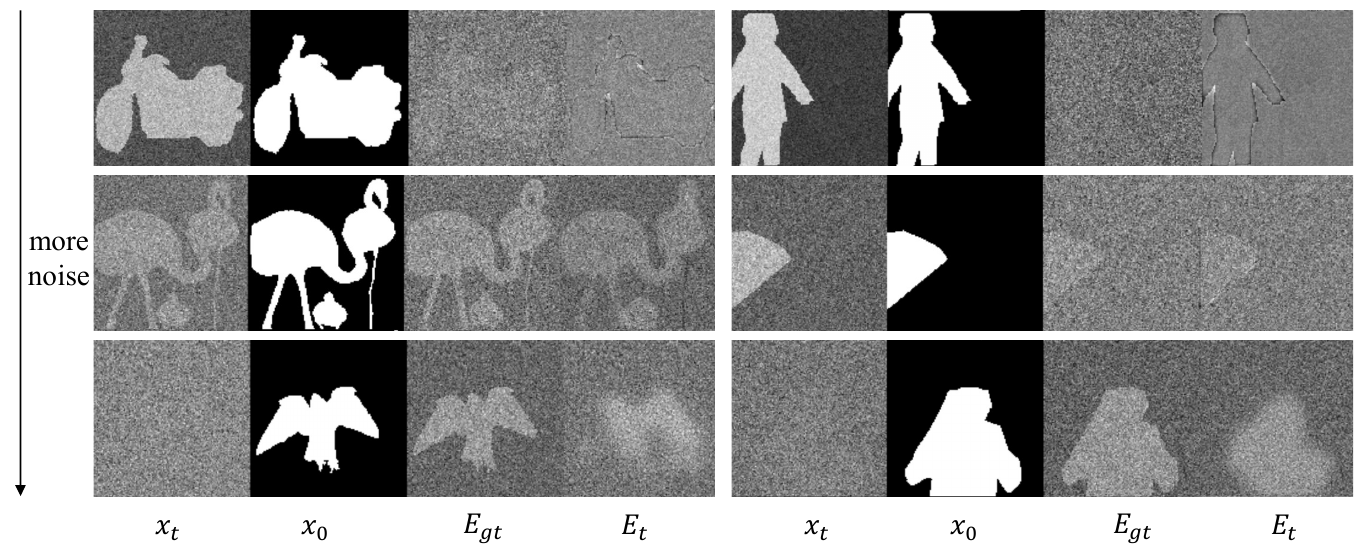}
    \caption{Visualization of error maps generated by the DQM module.}
    \label{vis:error-map}
  \end{figure}

  Fig.~\ref{vis:error-map} demonstrates our quality-aware module's exceptional noise robustness across low, medium, and high noise levels. Throughout the noise transition, the module consistently generates discriminative error maps that precisely locate problematic segmentation regions. This stability ensures reliable quality feedback even during highly degraded diffusion stages, providing spatially precise guidance that enables targeted improvements while preserving correct segmentations. These results validate our quality-aware strategy as an effective approach for dynamic segmentation optimization.

\section{Conclusion}
This paper introduces DiffUp, a novel FSS framework that addresses three critical challenges: seamless backbone upgrade without re-training, uniform handling of diverse annotations, and accommodation of diverse annotation quantities. This framework utilizes the proposed Backbone-Agnostic Feature Transformation (BAFT) to convert diverse annotations into backbone-agnostic multi-granularity priors, thereby facilitating seamless backbone upgrade. These multi-granularity priors are then aligned with the noisy intermediates in a diffusion process through the proposed Self-Conditioned Modulation block (SCM) and the Dual-Level Quality Modulation branch (DQM). Furthermore, an Uncertainty-Aware Prior Fusion (UAPF) block effectively harmonizes varying support quantities. Our approach not only advances the state-of-the-art in FSS but also offers a versatile solution for practical deployment.

\bibliographystyle{elsarticle-num} 
\bibliography{egbib}

\end{document}